\documentclass{article}

\usepackage{arxiv}
\usepackage{amsmath}
\usepackage[utf8]{inputenc} 
\usepackage[T1]{fontenc}    
\usepackage{hyperref}       
\usepackage{url}            
\usepackage{booktabs}       
\usepackage{amsfonts}       
\usepackage{nicefrac}       
\usepackage{microtype}      
\usepackage{graphicx}
\usepackage[labelsep=period]{caption}
\usepackage{amssymb}
\usepackage{natbib}

\title{Sparsity-driven emergence of functional modularity in multitask neural networks}

\author{Bowen Shi$^{1,2}$\begin{NoHyper}\footnotemark[1]\end{NoHyper} \quad Qi Ouyang$^{1,3,4}$ \quad Long Qian$^{2,3}$\begin{NoHyper}\footnotemark[2]\end{NoHyper}\\
\small $^1$Institute for Advanced Study in Physics, Zhejiang University, Hangzhou 310058, China\\
\small $^2$Center for Quantitative Biology, Academy for Advanced Interdisciplinary Studies,\\
\small Peking University, Beijing 100871, China\\
\small $^3$ZJU--Hangzhou Global Scientific and Technological Innovation Center, Hangzhou 311200, China\\
\small $^4$Institute of Fundamental and Transdisciplinary Research, Zhejiang University, Hangzhou, China\\
\texttt{bwshi1996@qq.com; qoy@zju.edu.cn; long.qian@pku.edu.cn}
}

\renewcommand{\today}{October 22, 2026}
\let\presubmaketitle\maketitle
\begin{document}

\maketitle

\begin{abstract}

Neural networks can learn multiple tasks, but how learning organizes shared computation into specialized and reusable components remains poorly understood. In this work, we show that sparsity-inducing regularization promotes functional modularity in multitask networks without predefined module boundaries. Using functional profiling in multilayer perceptrons and a lightweight transformer, we identified compact, task-selective subnetworks, termed functional tokens. Across rule-based task families, selectively retaining these tokens preserves their associated functions while reducing performance on other tasks. Such specialization develops preferentially in deeper layers during training, while earlier layers retain shared computation. Regularization strength and form shape module compactness and specialization, while task weighting alters functional allocation. Moreover, extracted functional tokens can be recombined through alternative pathways to reconstruct a learned composite computation without retraining, preserving their connectivity and learned weights. Our results suggest that sparsity-constrained optimization shapes not only parameter use but also the organization and reuse of learned computation.

\end{abstract}


\section{Introduction}

Deep neural networks have achieved wide success across perception, language, reasoning, and scientific discovery \cite{Bengio+chapter2007,Hinton06,goodfellow2016deep}. Despite substantial advances in architectural design \cite{heDeepResidualLearning2015,huangDenselyConnectedConvolutional2018} and optimization \cite{ruderOverviewGradientDescent2017}, the organizational principles that emerge during learning remain poorly understood. In particular, it is unclear how networks with distributed parameters and representations develop specialized functions while sharing and reusing computation across tasks. Biological intelligence provides a compelling example of such organization: functional modularity, characterized by partially independent yet interacting computational components, enables neural systems to support specialized processing, robustness, and compositional reuse across tasks \cite{damicelliTopologicalReinforcementPrinciple2019,margalitUnifyingFrameworkFunctional2024}.

Motivated by the computational advantages of modularity, previous machine-learning approaches have attempted to incorporate modular organization through explicit design. Neural module architectures introduce predefined functional components that can be selectively combined for different tasks \cite{andreasNeuralModuleNetworks2017,braylanReuseNeuralModules2015}, while adaptive structural learning methods seek to optimize network architectures by modifying connectivity patterns \cite{cortesAdaNetAdaptiveStructural2017,liuStructureLearningDeep2018}. Other approaches have explored structure-preserving constraints to incorporate prior organization into neural networks \cite{celledoniStructurepreservingDeepLearning2021}. More recently, sparse expert architectures have enabled large-scale neural networks to achieve conditional computation by routing inputs to specialized subnetworks \cite{shazeer2017outrageously,fedus2022switch}. Although these approaches demonstrate the utility of modular computation, they typically rely on architectural constraints, predefined components, or explicit routing strategies \cite{sunModularityDeepLearning2023}.

Modular organization, however, need not be specified in advance. Evolutionary studies have shown that modularly varying goals or selection to reduce connection costs can promote the emergence of modular networks \cite{kashtan2005spontaneous,clune2013evolutionary}. More recently, work on multitask recurrent networks has linked emergent modularity to task demands, representational capacity, and connection-growth dynamics \cite{wu2026taskstructured}. Learned modular connection masks can also improve subsequent learning after weight reinitialization, particularly under capacity constraints \cite{wu2026taskstructured}, demonstrating the utility of learned connectivity as a structural prior.

Sparsity provides an optimization-level route to investigate how constraints on connectivity shape computation. Sparsity-inducing regularization encourages a network to use fewer connections without prescribing internal module boundaries. Sparse and structured regularization have been extensively studied for model compression, parameter selection, and pruning \cite{wenLearningStructuredSparsity2016,scardapaneGroupSparseRegularization2017,maTransformed1Regularization2019}, while analyses of learned features, task-selective representations and shared dynamical motifs have revealed structure in trained networks \cite{agrawalWeightScaleInterpretingWeight2022,liGeometryConceptsSparse2024,michaudOpeningAIBlack2024,yangTaskRepresentationsNeural2019,driscollFlexibleMultitaskComputation2024}. The question here goes beyond network size: can sparsity help organize computation into task-selective components that preserve their associated functions and can be recombined with their learned parameters intact, without retraining?

In this work, we show that sparsity-inducing optimization can drive the emergence of functional modularity in multitask neural networks without explicitly designed modular architectures. Across multilayer perceptrons and a light-weight transformer, we identify compact, task-selective subnetworks, which we denote functional tokens, that coexist with shared computation. Across task families, these tokens preserve their associated functions under selective retention. By systematically varying sparsity strength and regularizer form, tracking functional specialization during training, and analyzing endpoint representation geometry, we characterize how sparsity shapes depth-dependent functional reorganization. Furthermore, extracted functional tokens can be connected through different pathways to reconstruct a learned composite computation without retraining or changing their learned parameter values. These findings connect sparse optimization to the organization and reuse of learned computation, highlighting machine-learning systems as models for uncovering general principles of computational organization \cite{levineMachineLearningMeets2024}.

\section{Results}

\subsection{Task-selective functional modules emerge under sparse multitask learning}

\begin{figure}[p]
    \centering
    \includegraphics[width=\textwidth,height=0.79\textheight,keepaspectratio]{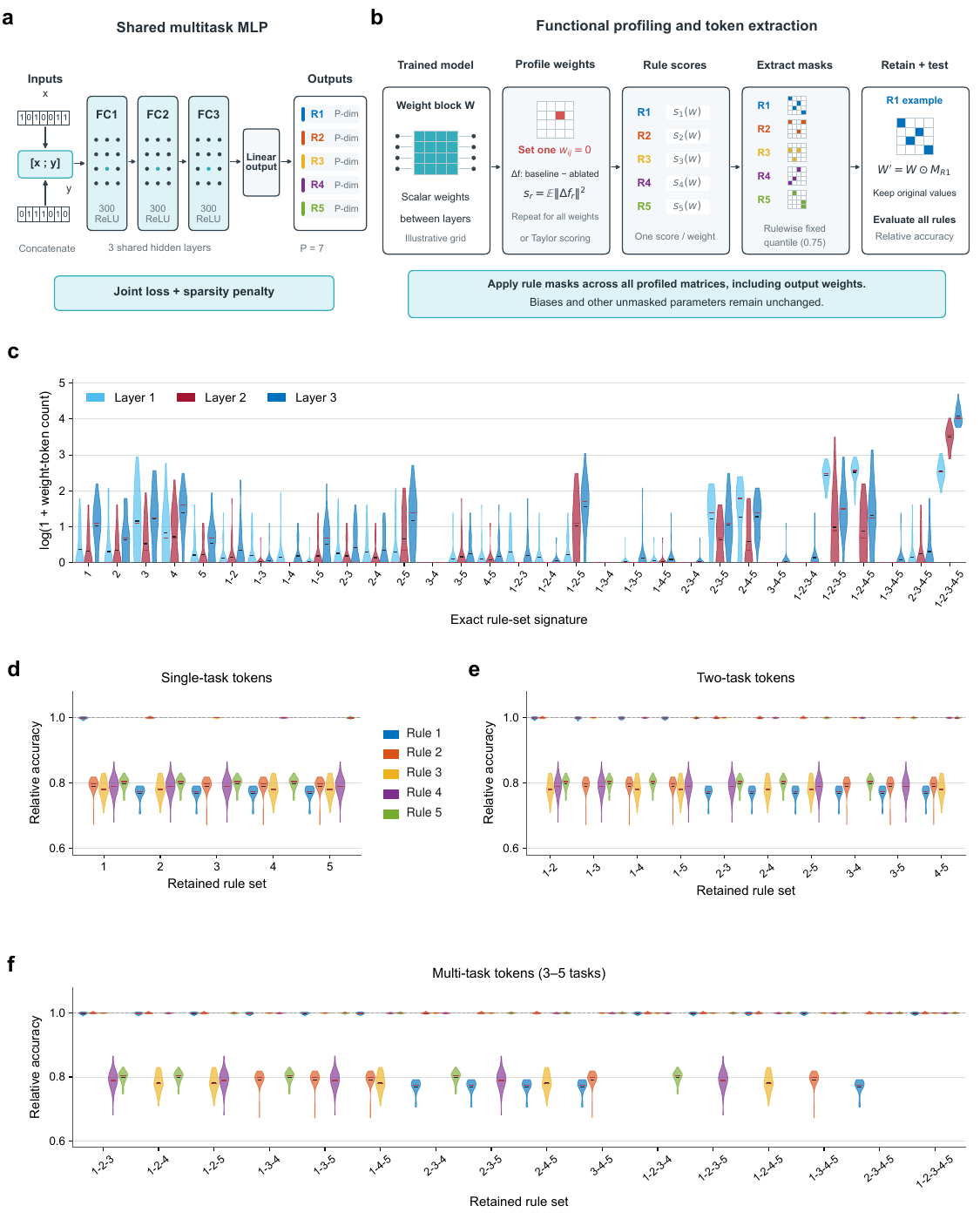}
    \caption{Network architecture, functional profiling and task-selective weight supports. (a) Shared MLP with three 300-unit fully connected (FC) ReLU hidden layers, trained jointly on five rules (R1--R5 of Binary Sequence Calculation rules). (b) Functional profiling and extraction. Each scalar weight is scored individually by output ablation or Taylor output-sensitivity profiling. Rule-specific masks are constructed separately for each profiled weight matrix and then applied simultaneously across those matrices, including the output weights; biases and parameters outside the masking set are unchanged. (c) Log-transformed weight counts, $\log(1+\mathrm{count})$, for the 31 nonempty exact rule-set signatures, shown separately for the three hidden-layer weight matrices. For example, ``1--3'' denotes the weights selected for exactly rules 1 and 3. (d--f) Relative accuracy of subnetworks retaining the union of masks for one rule (d), two rules (e), or three to five rules (f). Colours identify the evaluated rule. Relative accuracy is retained-subnetwork bitwise accuracy divided by that of the full network on the same rule. Violins summarize 30 independent runs; black and red marks indicate means and medians, respectively.}
\label{fig:main1}
\end{figure}

To examine how computation is shared and specialized across tasks, we used a controlled setting where a multilayer perceptron (MLP) with three fully connected ReLU hidden layers was trained to perform five related Binary Sequence Calculation tasks, without task-specific hidden branches or explicit modular routing (Fig.~\ref{fig:main1}a, Methods). The task set comprised four primitive arithmetic transformations, \(a + b\), \(ab\), \(a^{2}\), and \(b^{2}\), and one composite transformation \(a(a + b)\). Each shared input pair carried targets for all five rules, whose predictions occupied separate output blocks. The network was trained jointly on all rule losses from random initialization under sparsity-inducing \(L_{\gamma}\) regularization, with \(\gamma = 1\) as the default (Methods; Supplementary Note 5.1).

To map functional contributions across tasks, we individually silenced scalar weights and measured the resulting change in each rule's output segment (Fig.~\ref{fig:main1}b). Each scalar weight served as an elementary token. Rule-specific masks were constructed at a fixed extraction quantile, $\tau=0.75$ (Methods). Each selected weight was assigned an exact rule-set signature specifying the rules for which it met this criterion. We define a functional token for a rule or rule set by retaining the weights selected by the corresponding mask or union of masks while silencing the remaining weights simultaneously across all profiled weight matrices, including the output weights. Biases and other parameters outside the masking set remained unchanged (Methods; Supplementary Notes~\ref{appendix:Taylor} and~\ref{appendix:implementation}).

The resulting weight assignments revealed a depth-dependent organization (Fig.~\ref{fig:main1}c). Earlier hidden-layer weight blocks showed broader multi-rule support, whereas deeper blocks were enriched in sharper single-rule signatures and signatures involving fewer rules. This pattern was reproducible across independently trained networks.

We next examined whether retaining these supports preserved their associated computations. Rule-wise relative accuracy was defined as the retained subnetwork's accuracy divided by that of the corresponding full network on the same task. Single-rule functional tokens preserved substantial relative accuracy on their target rules, while performance on the other rules was reduced (Fig.~\ref{fig:main1}d). This selective preservation therefore provided a functional test of the extracted supports distinct from the individual-weight output perturbations used to define them. Multi-rule functional tokens likewise retained performance on their constituent tasks while remaining less effective on rules outside the selected rule set (Fig.~\ref{fig:main1}e,f).

\subsection{Functional modularity persists across task families, profiling methods, and model architectures}

Having identified task-selective functional modules in the Binary Sequence Calculation setting, we next asked whether this organization depended on task semantics or input format, profiling criterion, or model architecture.

\begin{figure}[p]
\centering
\includegraphics[width=\textwidth,height=0.82\textheight,keepaspectratio]{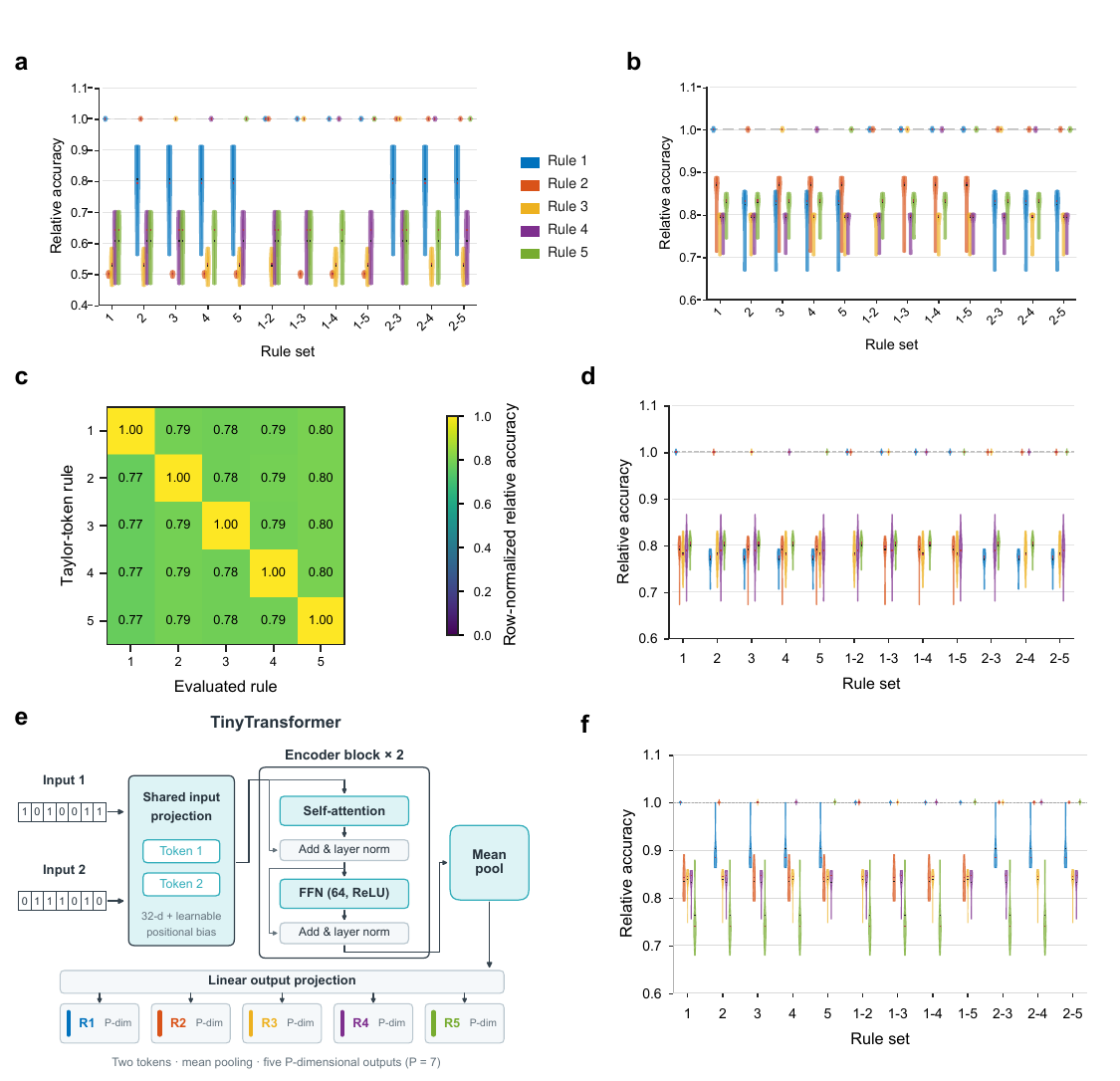}
\caption{Functional-token selectivity across task families, profiling methods and architectures. (a,b) Relative accuracy of ablation-derived retained masks for Sparse Pattern Logic (a) and MNIST Digit-Pair Arithmetic (b), respectively. The Binary Sequence Calculation baseline is shown in Fig.~\ref{fig:main1}d--f. (c) Mean relative accuracy matrix for single-rule Taylor-derived tokens from 30 independent MLP runs, with each retained-token row normalized to its maximum. Rows identify the retained-token rule and columns the evaluated rule. (d) Relative accuracy of Taylor-derived MLP subnetworks in Binary Sequence Calculation. (e) TinyTransformer architecture with two input-sequence tokens, a shared 32-dimensional projection and positional bias, two self-attention/FFN blocks with residual connections and layer normalization, mean pooling and a linear output head. (f) Relative accuracy of Taylor-derived TinyTransformer subnetworks in Binary Sequence Calculation. In a,b,d,f, the horizontal axis identifies the retained rule set, colors identify the evaluated rule, and accuracy is normalized to the corresponding full network on that rule. Violins summarize 30 independent runs in a,b,d and 23 batch-128 runs in f; black and red marks indicate means and medians, respectively.}
\label{fig:main3}
\end{figure}

We first trained the same shared MLP architecture on two additional task families. Sparse Pattern Logic replaced arithmetic transformations with five Boolean-style rules, whereas MNIST Digit-Pair Arithmetic retained the arithmetic rule structure but presented pairs of digit images as inputs, coupling image-based input processing to arithmetic prediction (Methods; Supplementary Notes 5.2 and 5.3). Across independently trained networks for each task family, the extracted functional tokens preserved higher relative accuracy on their associated rules than on omitted rules (Fig.~\ref{fig:main3}a,b; Binary Sequence Calculation baseline in Fig.~\ref{fig:main1}d--f; Methods). Residual performance on omitted rules varied across task families, reflecting differences in computational overlap among the underlying transformations. Thus, the extracted subnetworks captured both task-selective and shared structure.

We next replaced exhaustive output ablation with randomized Taylor output-sensitivity profiling, a computationally less expensive criterion that estimates the squared first-order output perturbation associated with removing a scalar weight (Methods; Supplementary Note~\ref{appendix:Taylor}). The row-normalized mean relative-accuracy matrix for single-rule Taylor-derived tokens peaked on the corresponding evaluated rules (Fig.~\ref{fig:main3}c). Functional tokens extracted using the same score-quantile criterion retained high relative accuracy on their associated rules and showed reduced performance on other rules (Fig.~\ref{fig:main3}d). Thus, the two profiling criteria recovered consistent patterns of target-task sufficiency and off-task selectivity, indicating that the observed modular organization was not specific to a single importance measure.

Finally, we trained a TinyTransformer on the same five-rule Binary Sequence Calculation task set (Fig.~\ref{fig:main3}e; Methods; Supplementary Note 2). Applying the Taylor-based profiling and retention procedure to scalar weight coordinates yielded functional tokens that preserved the corresponding target functions, with lower performance on omitted rules (Fig.~\ref{fig:main3}f). The extent of performance reductions varied across tasks. This architectural control extended the observation of task-selective functional organization beyond fully connected MLPs.

\subsection{Sparse regularization shapes the compactness and specialization of functional supports}

Having shown that task-selective functional organization persists across the tested settings, we next asked whether sparsity merely accompanies modularity or actively controls its emergence. Specifically, we interrogated the effects of the regularization strength $l$ and the regularization exponent $\gamma$ in the multitask MLP.

\begin{figure}[p]
\centering
\includegraphics[width=\textwidth,height=0.78\textheight,keepaspectratio]{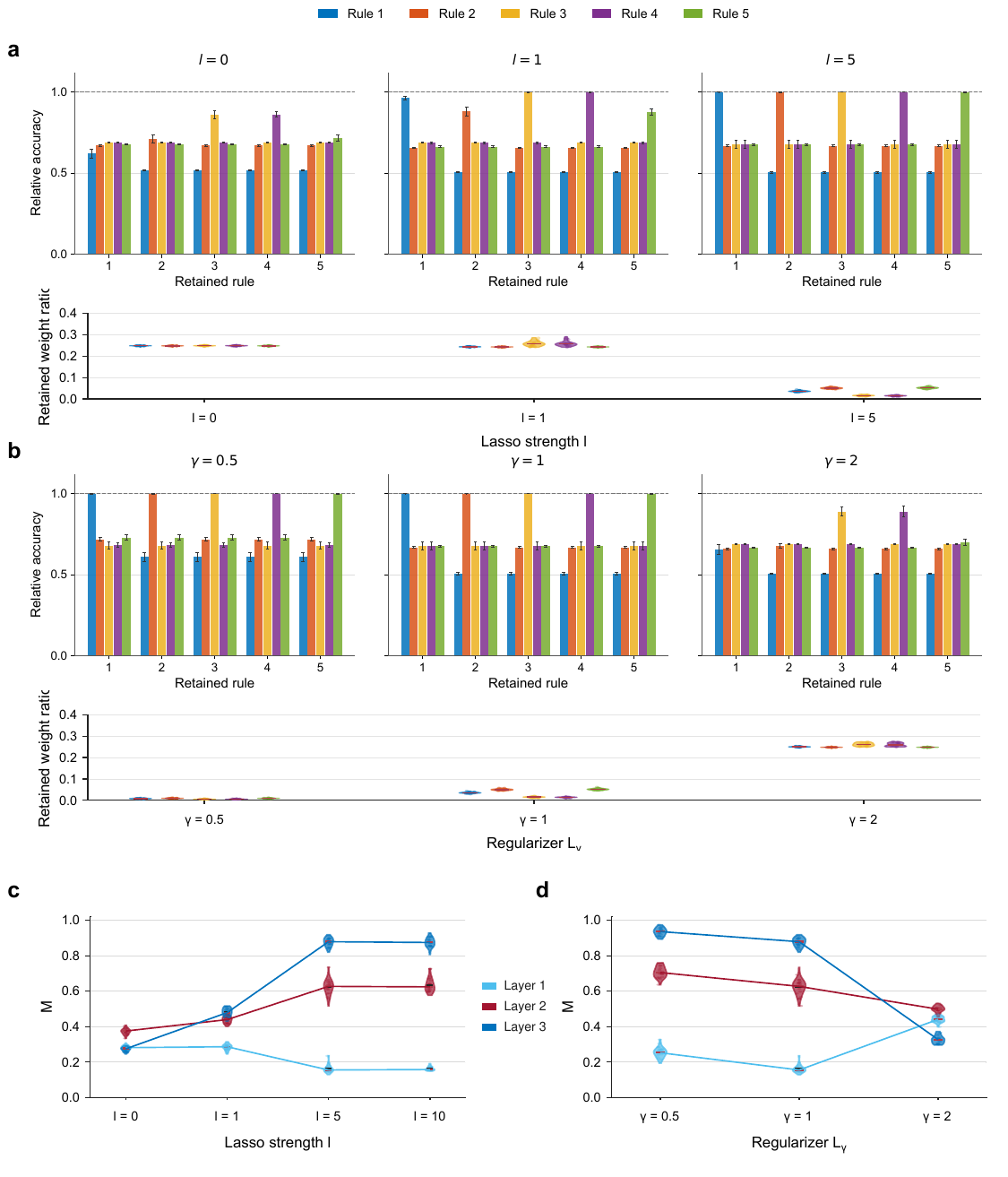}
\caption{Regularization changes the compactness and selectivity of extracted weight supports. (a) Relative accuracy of single-rule subnetworks at Lasso strengths $l=0,1,5$ with $\gamma=1$ (upper plots), and their retained-weight ratios (lower plot). (b) The corresponding quantities for regularizer exponents $\gamma=0.5,1,2$ at $l=5$. Bar colors identify the evaluated rule; colors in the lower plots identify the rule defining each retained functional token. Bars show means and error bars show sample SD across 30 independent runs. The retained-weight ratio is the number of mask-selected scalar coordinates divided by the total number of scalar coordinates in the analyzed weight matrices. The extraction quantile is fixed at $\tau=0.75$; score thresholds are determined separately for each network, matrix and rule. (c,d) Layer-wise modularity metric across Lasso strengths $l=0,1,5,10$ with $\gamma=1$ (c) and across regularizer exponents $\gamma=0.5,1,2$ at $l=5$ (d). Colored guide lines connect medians for each layer at aligned conditions (x positions). All panels use ablation profiling and 30 independent runs per condition. Panels c,d show importance-weighted normalized inverse participation (NIP), defined in Eq.~\eqref{eq:modularity}. The analysis uses $\epsilon_S=10^{-9}$, $\epsilon_{\mathrm{out}}=10^{-24}$ and all scalar-weight candidates (Methods).}
\label{fig:main4}
\end{figure}

We first compared Lasso regularization strength \(l\) (\(\gamma = 1\)) with the non-regularized control (\(l = 0\)), re-extracting functional tokens from each trained network using the same ablation-based, within-rule score quantile criterion. For each \(l\) condition, three complementary quantities were assessed: target-rule relative accuracy, the retained scalar weight fraction, and the importance-weighted modularity statistic $M$ (Eq.~\eqref{eq:modularity}; Fig.~\ref{fig:main4}; Methods).

Without sparsity regularization, the extracted functional tokens retained comparatively large and diffuse supports and showed incomplete preservation of their target functions. Introducing \(L_{1}\) regularization at \(l = 1\) substantially improved target-rule relative accuracy while retained-weight fractions remained similar to those in the unregularized condition. Increasing the regularization strength to \(l = 5\) markedly reduced these fractions, with target-rule relative accuracy remaining close to the full-network level (Fig.~\ref{fig:main4}a). At $l=0$, the retained-weight ratio was close to one quarter, primarily reflecting the 75th-percentile extraction rule.

The modularity score $M$ provided a complementary measure of how selectively these weights contributed to different rules. $M$ incorporates the magnitude of the profiled contributions and measures their concentrations on individual rules (importance-weighted normalized inverse participation, NIP; Methods; Eq.~\eqref{eq:modularity}). Increasing $l$ from 0 to 5 raised $M$ in the second and third hidden-layer weight blocks, followed by a plateau at $l=10$ (Fig.~\ref{fig:main4}c). The first block did not show the same monotonic response. Higher $M$ reflected more rule-concentrated importance profiles among the selected weights (Supplementary Note~\ref{appendix:Taylor}).

We next asked whether these effects depended on the form of the regularizer. At fixed $l=5$, we compared ${L}_{\gamma }$ penalties with $\gamma \in \left\{0.5,1,2\right\}$ (Methods; Supplementary Note 2). The ${L}_{0.5}$ and ${L}_{1}$ penalties produced compact functional tokens that preserved target-rule performance and showed stronger deeper-block specialization, whereas the ${L}_{2}$ penalty yielded substantially larger retained-weight fractions, lower target-rule relative accuracy, and lower $M$ in deeper blocks (Fig.~\ref{fig:main4}b,d).

A local support-recovery analysis provides a conditional framework for interpreting both the dependence on regularizer form and the layer-specific response (Supplementary Note 1). Approximating hidden-layer computation or parameter-Jacobian effects by a multitask linear surrogate, the analysis identifies conditions on signal strength, feature coherence, and local approximation error under which sparse task supports can be recovered or locally stabilized. For $\gamma =1$, support recovery follows from a Lasso primal-dual witness argument; for $0<\gamma <1$, the nonconvex objective can admit an oracle-supported strict local minimum. In contrast, penalties with $\gamma >1$ primarily shrink weight magnitudes without enforcing exact sparsity \cite{zhaoModelSelectionConsistency,maTransformed1Regularization2019}. This distinction offers an interpretation of the contrast in Fig.~\ref{fig:main4}b,d: sparsity-inducing penalties can stabilize sparse task supports once compatible local coordinates emerge.

Theoretical analyses further indicate that support recovery or stabilization for \(0 < \gamma \leq 1\) depends critically on the coherence structure of the effective design, particularly the off-diagonal structure of the feature covariance. When active and inactive coordinates are sufficiently distinguishable, sparse regularization can more effectively eliminate redundant coefficients; strongly correlated features make exact support separation less stable. This dependence offers a possible interpretation of the observed depth profile. Early layer weights operate on common inputs and may retain intermediate features shared across tasks. Successive nonlinear transformations can produce coordinates in which task-relevant effects are more readily separated, allowing sparse penalties to concentrate rule-selective influence in later blocks closer to the rule-conditioned outputs while preserving shared computation near the input.

Together, retained-weight fraction, relative accuracy, and \(M\) distinguish support compactness, functional preservation, and specialization. The strength and form of sparse regularization shape how task-relevant influence is concentrated into compact, task-selective supports, while allowing shared computation to persist in earlier layer transformations.

\subsection{Depth-dependent functional specialization develops during training}

\begin{figure}[p]
\centering
\includegraphics[width=\textwidth,height=0.82\textheight,keepaspectratio]{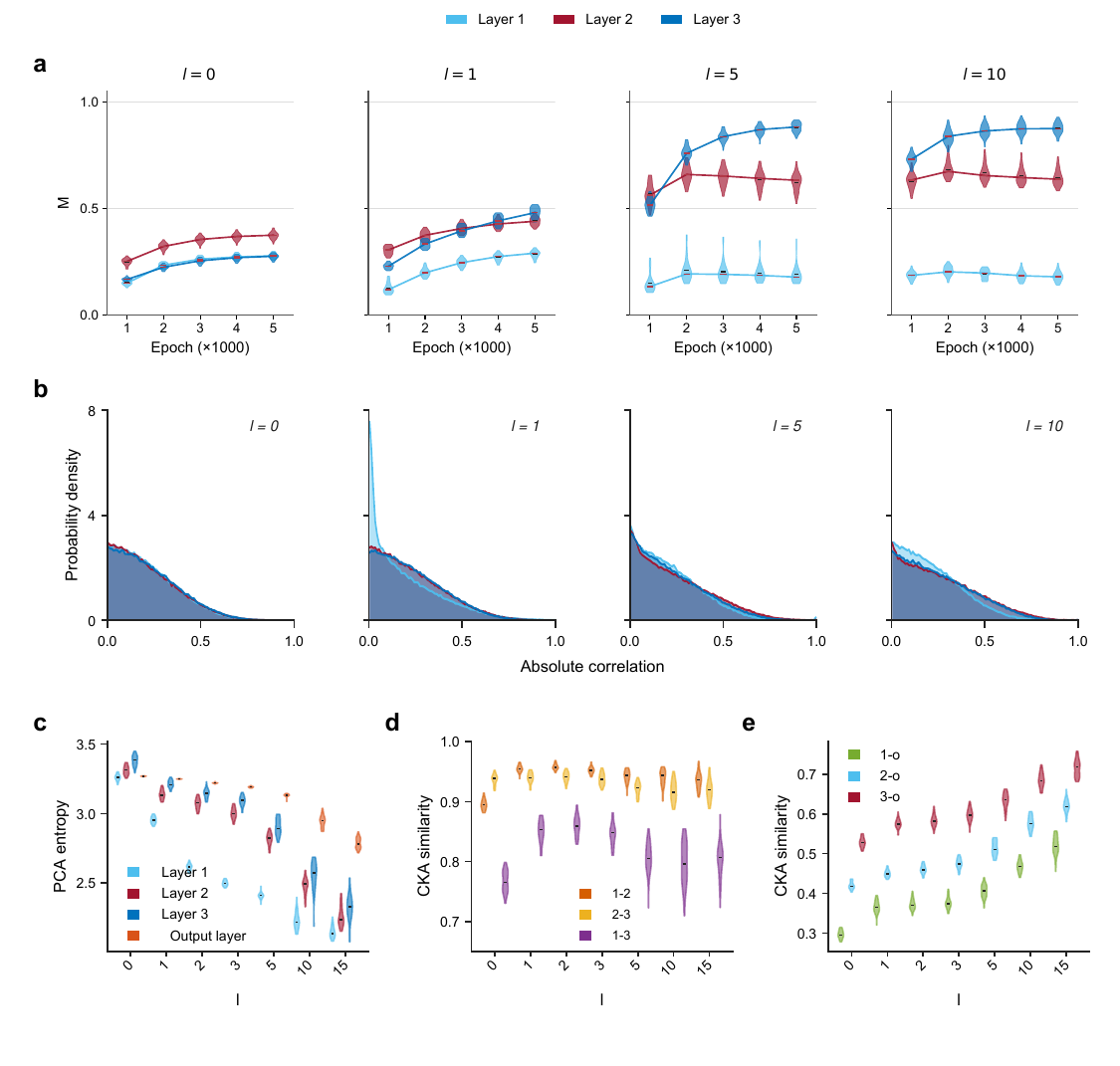}
\caption{Training dynamics and endpoint representation geometry. (a) Importance-weighted modularity $M$ at 1,000--5,000 epochs for $l=0,1,5,10$, from left to right, with $\gamma=1$. Masks are recomputed at each checkpoint using Taylor output-sensitivity profiling with four probes and the same extraction quantile, $\tau=0.75$. Violins show 30 independent runs; black and red marks show means and medians. Coloured lines connect layer-wise medians at common epochs. The analysis uses $\epsilon_S=10^{-9}$, $\epsilon_{\mathrm{out}}=10^{-8}$, seed 0 and all scalar-weight candidates (Methods). (b) Distributions of absolute pre-activation unit correlations at the same four Lasso strengths, aligned column by column to (a) and pooled from four runs per condition. (c) PCA spectral entropy of post-ReLU hidden-layer activations and output representations. (d,e) Linear centered kernel alignment (CKA) between hidden-layer pairs (d) and between hidden and output layers (e). Endpoint metrics in c--e use 30 runs per condition over $l=0,1,2,3,5,10,15$. Hidden-layer colours are shared across a-c; colors in d,e identify layer pairs, with ``o'' denoting the output layer.}
\label{fig:main5}
\end{figure}

To determine how task-selective functional influence developed during optimization, we tracked the importance-weighted modularity score \(M\) at five saved checkpoints spanning 1,000--5,000 training epochs under Lasso strengths \(l \in \{ 0,1,5,10\}\) (Fig.~\ref{fig:main5}a). At each snapshot, scalar weights were Taylor-profiled and masks were recomputed within each hidden-layer weight block using the same rule-wise score-quantile criterion (\(\tau = 0.75\)). Each condition comprised 30 independent runs.

The trajectories were strongly dependent on both depth and regularization strength. Without regularization, median $M$ increased only modestly and reached 0.278, 0.375, and 0.277 in the three blocks at epoch 5000. At $l=1$, specialization accumulated more clearly in the second and third blocks, whose final medians were 0.440 and 0.481. Stronger regularization produced a more pronounced increase in the deepest block: between epochs 1,000 and 5,000, median $M$ rose from 0.516 to 0.883 at $l=5$, and from 0.731 to 0.874 at $l=10$. In contrast, the second block attained intermediate values and approximately plateaued, whereas the first block remained comparatively weakly specialized. Thus, task-selective importance progressively concentrated in the deepest block rather than increasing uniformly across the network. This layer-dependent temporal separation mirrors the final depth-dependent organization: early transformations retained broader multi-rule influence, whereas the deepest weight block progressively consolidated task-selective importance.

To characterize the representations accompanying this organization, we analyzed the latest saved epoch with \(\gamma = 1\) (Fig.~\ref{fig:main5}b--e; Supplementary Note 6). Pairwise-correlation distributions were estimated at \(l \in \{ 0,1,5,10\}\) from four independent runs per condition, whereas PCA and centered kernel alignment (CKA) analyses used 30 runs per condition over the extended sweep \(l \in \{ 0,1,2,3,5,10,15\}\).

Pearson correlations between pre-activation unit responses changed in a layer-dependent, non-monotonic manner (Fig.~\ref{fig:main5}b). The clearest increase in correlations near zero occurred in the first hidden layer at \(l = 1\). At stronger regularization, deeper layers instead redistributed more correlation mass toward intermediate values.

 PCA spectral entropy revealed a complementary change in the distribution of representational variance (Fig.~\ref{fig:main5}c). Across the analyzed post-ReLU hidden and output representations, the pooled median entropy decreased from 3.279 at \(l = 0\) to 2.280 at \(l = 15\), indicating that representational variance became concentrated into fewer dominant directions. Linear CKA further showed that sparsity altered the division of representational roles across layers (Fig.~\ref{fig:main5}d,e). Similarity among hidden-layer representations remained high across the sweep, peaked at moderate regularization, and declined modestly at larger $l$. By contrast, pooled hidden-output CKA increased from 0.418 at $l=0$ to 0.619 at $l=15$. Hidden representations therefore retained broadly compatible geometry while becoming progressively aligned with the rule-conditioned output space.

Together, the training trajectories show that sparse optimization preferentially strengthens task-selective influence in deeper weight blocks. Endpoint representation analyses associate this organization with concentrated variance and stronger hidden-output alignment, rather than uniform decorrelation across layers. These observations are compatible with the support-recovery interpretation (Supplementary Note 1), by which sparse task supports are most readily stabilized when active and inactive coordinates are sufficiently separable, which depends on the coherence of the effective design matrix as well as on signal strength and local approximation error.

\subsection{Task-biased optimization reshapes multitask performance and functional allocation}

\begin{figure}[p]
\centering
\includegraphics[width=\textwidth,height=0.71\textheight,keepaspectratio]{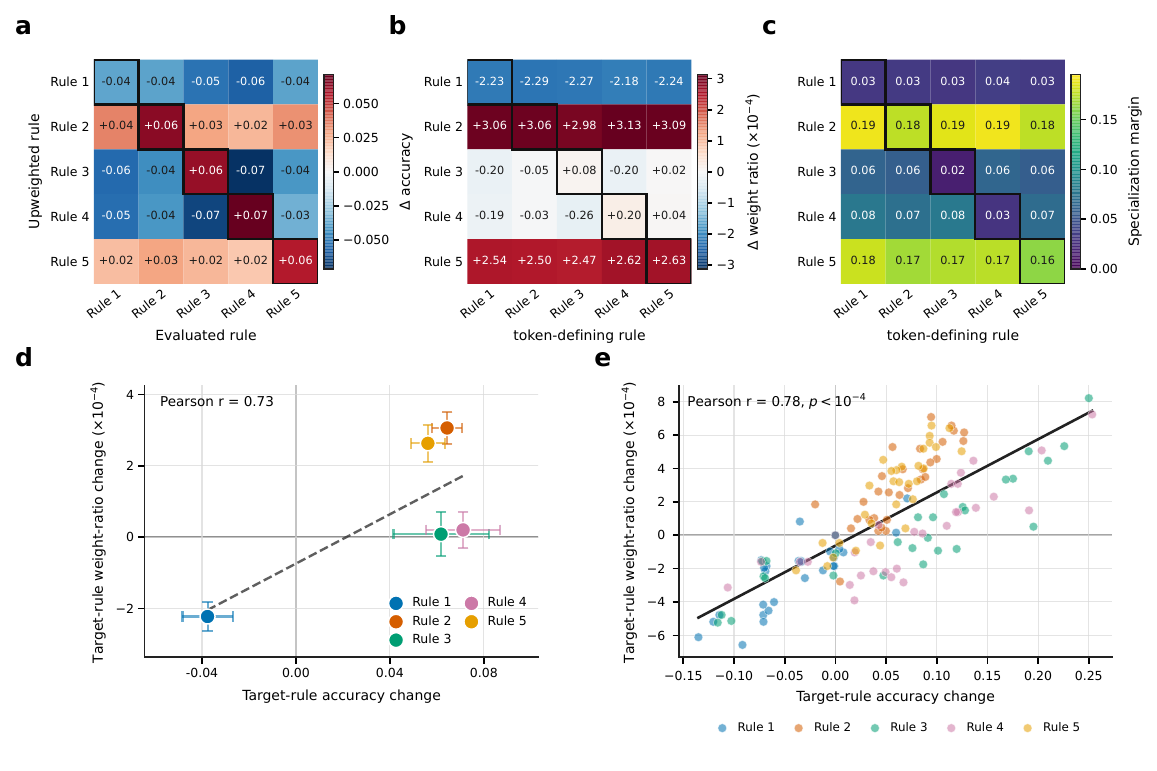}
\caption{Rule-loss weighting changes network performance and functional-token allocation. Each intervention assigns five-fold loss weight to one rule and unit weights to others, and is compared with an equal-weight control, with 30 independent runs per condition at $l=5$ and $\gamma=1$. Rows in a--c identify the upweighted rule; columns identify the evaluated rule in a and the token-defining rule in b,c. (a) Mean change in full-network accuracy for each upweighted-rule and evaluated-rule pair, calculated as intervention minus control. (b) Mean change in the retained scalar-weight fraction, in units of $10^{-4}$. (c) Mean specialization margin: relative accuracy on the token-defining rule minus its mean relative accuracy on the other four rules. Outlined diagonal cells in a--c identify the upweighted target. (d) Association between the five target-rule mean accuracy changes and retained-weight-ratio changes; error bars combine intervention and control marginal uncertainties as defined in Supplementary Note~7. The dashed line denotes a least-squares fit. (e) Association across 150 repeat-matched target-rule responses, with 30 responses per upweighted rule. Colors in d,e denote the upweighted rule. The solid line in e is a least-squares fit and the annotation reports the pooled Pearson correlation and nominal two-sided $p$ value.}
\label{fig:main6}
\end{figure}

Having examined how sparse regularization shapes functional organization, we next asked whether preferentially optimizing one task under sparse regularization improves its performance at the expense of other tasks, and how these changes relate to functional specialization. Without changing the network architecture or training data, we introduced per-rule weighting into the multitask objective:

\[\mathcal{L}=\frac{1}{R}\sum_r W_r\mathcal{L}_r+l\Omega_\gamma(\theta).\]

where ${W}_{r}$ denotes the weight assigned to rule $r$. To emphasize a specific rule, we assigned ${W}_{r}=5$ while keeping unit weights for all other rules. All six conditions used three hidden layers of 300 units and were trained with $l=5$ and $\gamma=1$, and each weighting condition was compared with an equal-weight control across 30 runs (Fig.~\ref{fig:main6}).

The full-network response depended on the emphasized rule (Fig.~\ref{fig:main6}a). Upweighting rules 2 or 5 improved mean accuracy not only on the target rule but also on all four non-upweighted rules. By contrast, emphasizing rules 3 or 4 improved target-rule accuracy while reducing mean accuracy on the other rules. Upweighting rule 1 reduced performance on both the target and non-target rules. Preferential task weighting therefore produced distinct outcomes, ranging from cross-task gains to target-specific gains accompanied by losses elsewhere.

To relate these performance changes to functional organization, we re-extracted functional tokens independently from each trained network using the same ablation-based score-quantile criterion. We quantified each token by its retained scalar weight fraction and its specialization margin, defined as its target-rule relative accuracy minus its mean relative accuracy on the other four rules (Supplementary Note 7). Support changes extended across rule-defined tokens: emphasizing rules 2 or 5 increased retained-weight fractions broadly, whereas emphasizing rule 1 reduced them. Rules 3 and 4 produced small increases in their own target-token supports, with mixed increases and decreases in non-target supports (Fig.~\ref{fig:main6}b). Thus, the observed performance trade-offs did not simply correspond to expansions of the target token accompanied by the contraction of the others.

Specialization margins further distinguished functional selectivity from preferential task performance. Conditions emphasizing rules 2 and 5 yielded margins of approximately 0.16--0.19 across the rule-defined tokens, whereas margins were substantially smaller under the rule-3 and rule-4 conditions that exhibited cross-task performance costs (Fig.~\ref{fig:main6}c).

Changes in target accuracy and retained-weight fraction were positively associated across the five target-rule means (Pearson \(r=0.73\); Fig.~\ref{fig:main6}d) and across 150 repeat-matched target-rule responses (\(r=0.78\), nominal \(p<10^{-4}\); Fig.~\ref{fig:main6}e). Target-performance gains therefore tended to accompany larger selected supports, but were not restricted to conditions with high functional selectivity. The pooled association comprises five upweighted-rule conditions with 30 repeat-matched responses per condition (Supplementary Note~7).

\subsection{Learned functional tokens support compositional reuse without retraining}

Finally, we tested whether the extracted functional tokens could reuse their learned computations in explicitly specified composite pathways outside the configuration in which they were originally identified.

\begin{figure}[p]
\centering
\includegraphics[width=\textwidth,height=0.71\textheight,keepaspectratio]{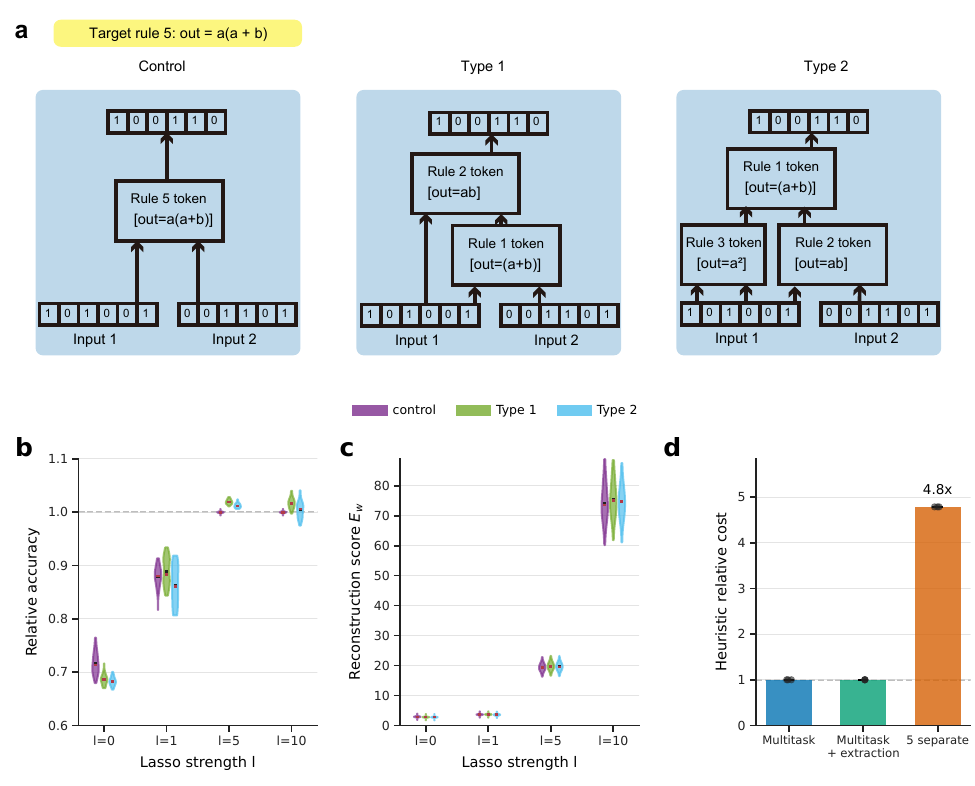}
\caption{Recomposition of learned functional tokens without retraining. (a) Recomposition of rule 5 computation $a(a+b)$. The directly extracted rule-5 token is the control; Type 1 computes $a(a+b)$ sequentially, and Type 2 computes $a^2+ab$ through parallel primitives followed by addition. Component outputs are thresholded and encoded using the original task interface. Component masks and learned weights remain fixed. (b) Rule-5 relative accuracy normalized to that of the full mother network. (c) Reconstruction score $E_w=A(C,5)/\rho_{\mathrm{nz},5}(l)$, where $A(C,5)$ is the relative accuracy for configuration $C$, and $\rho_{\mathrm{nz},5}(l)$ is the nonzero scalar-weight fraction of the directly extracted rule-5 token. The denominator is shared across all three configurations. In b,c, $l$ denotes the mother network Lasso strength; violins summarize 30 independent runs per regularization setting, with means and medians marked in black and red. (d) Heuristic training/extraction cost comparison, normalized within each run to multitask training plus extraction. Bars show means across three benchmark runs, error bars show s.e.m., and points show individual runs. This proxy excludes recomposition (Supplementary Note~8.7).}
\label{fig:main7}
\end{figure}

We considered the rule-5 Binary Sequence Calculation transformation,

\[f_5(a,b)=a(a+b)=a^2+ab.\]

which admits two equivalent decompositions through addition, multiplication, and squaring operations already learned by the same multitask network (Fig.~\ref{fig:main7}a).

For each Lasso strength \(l\), rule-specific functional tokens were extracted from a common mother network. Each component retained its extraction mask and learned parameters throughout recomposition (Supplementary Note 8). The directly extracted rule-5 token served as a non-recomposed control, with performance reported relative to the corresponding full mother network on rule 5.

In the type 1 pathway, the addition token first computed \(a + b\). Its thresholded output was paired with the original input \(a\) and passed to the multiplication token to implement \(a(a + b)\). In the type 2 pathway, the square and multiplication tokens operated in parallel to compute \(a^{2}\) and \(ab\); their thresholded outputs were then passed to the addition token to implement \(a^{2} + ab\). Thus, the two pathways reconstructed the same target task through distinct computational graphs. Both the sequential type 1 pathway and the parallel type 2 pathway recovered rule 5 with substantial relative accuracy (Fig.~\ref{fig:main7}b). Type 1 had a slightly higher mean relative accuracy than Type 2 at all four displayed regularization settings. At $l=5$ and $l=10$, both reached or exceeded full mother-network performance. These results showed that rule-associated components could preserve useful computations after being connected through the specified intermediate interfaces.

Recomposition was also possible without regularization, but the associated functional supports were substantially larger and more diffuse. To relate reconstruction accuracy to support compactness, we defined the parameter-normalized reconstruction score \(E_{w}\):

\[E_w=\frac{A_{C,5}}{\rho_{\mathrm{nz},5}(l)}.\]

where $A_{C,5}$ is relative accuracy on rule 5 and $\rho_{\mathrm{nz},5}(l)=N_{\mathrm{nonzero},5}/N_{\mathrm{total}}$ is the nonzero scalar-weight fraction of the directly extracted rule-5 control (Supplementary Note~\ref{appendix:recomposition}). The same denominator is used for the control and both recomposition pathways. $E_w$ increased with $l$ (Fig.~\ref{fig:main7}c), reflecting the combination of preserved reconstruction accuracy and a smaller reference support.

To assess the cost of obtaining functional tokens, we compared multitask training with and without token extraction against training five separate task-specific networks (Fig.~\ref{fig:main7}d; Supplementary Note~8.7). Across three auxiliary benchmark runs, the separate-training heuristic total was 4.78 times the joint-training-plus-extraction total.

\clearpage
\section{Discussion}

Our results demonstrate that functional modularity can emerge under sparsity-constrained multitask learning, with compact, task-selective supports coexisting with shared computation. This organization recurs across the tested task families, profiling criteria, and architectures, while the MLP analyses reveal a systematic depth dependence. These findings complement work on explicit modular architectural design \cite{andreasNeuralModuleNetworks2017,sunModularityDeepLearning2023} and task-driven modular organization \cite{wu2026taskstructured} by identifying sparse optimization as a control of functional allocation and testing the preservation and reuse of learned computation.

Our analyses distinguish support compactness, functional preservation and specialization. Under the tested conditions, sparsity-inducing penalties produced smaller and more selectively organized supports than \(L_{2}\) regularization. Drawing on insights from Lasso consistency \cite{zhaoModelSelectionConsistency}, the local support-recovery analysis distinguishes sparse support formation from magnitude shrinkage and identifies conditions on signal strength, feature coherence, and approximation error under which recovery or local stabilization is possible (Supplementary Note~1). Its dependence on effective coordinates offers a possible interpretation of the depth profile: nonlinear transformations may make task-relevant effects more readily separable in later blocks, while early transformations retain shared computation.

Training trajectories show that task-selective importance progressively concentrates in deeper blocks rather than increasing uniformly across the network. Separately, endpoint representation analyses associate this organization with concentrated variance and stronger hidden--output alignment, alongside layer-dependent, non-monotonic changes in neuronal correlations. These observations connect functional allocation to questions of feature geometry \cite{liGeometryConceptsSparse2024} and learned computational structure \cite{cagnettaTheoryHowStructure2024,chughtaiToyModelUniversality2023}, and resonate with neuroscientific accounts of modular specialization \cite{damicelliTopologicalReinforcementPrinciple2019}.

Task-biased optimization further distinguishes preferential task performance from functional specialization. Conditions with cross-task gains exhibited higher specialization margins than those with target--non-target performance trade-offs. Support changes also extended beyond the emphasized token rather than consistently following a target-expansion, non-target-contraction pattern. Cross-task costs therefore cannot simply be attributed to induced over-specialization. Instead, these results motivate evaluating functional selectivity jointly with performance across the task set, rather than treating greater task priority as a direct means of creating more isolated modules.

Methodologically, the functional-token framework connects importance profiling with tests of computational utility, complementing attribution-based interpretability approaches \cite{marcinkevicsInterpretableExplainableMachine2023,michaudOpeningAIBlack2024}. Retaining selected weights while simultaneously masking all profiled weight matrices, including output weights, preserved substantial target function and reduced off-target performance; biases and other unmasked parameters remained unchanged. Compositional reuse provides a further test: extracted functional tokens from a common mother network reconstruct a learned composite rule without retraining while preserving their masks and inherited weights. This evaluates reuse of learned computation, distinct from testing the learning performance of subnetworks trained in isolation under modular connection masks \cite{wu2026taskstructured}.

These recomposition experiments support the functional reuse of the identified tokens. Within the controlled task setting examined here, these properties support interpreting functional tokens as optimization-dependent and intervention-validated candidate sub-computations, while also showing that their reuse remains constrained by intermediate-interface compatibility and the selected computational decomposition.

Several limitations define the scope of this interpretation. Rule boundaries, primitive operations, composite relations, and task-specific readouts are specified by design, leaving open whether comparable supports can be recovered when tasks are implicit or component roles depend on context. Extraction depends on the intervention target, score normalization, and mask-selection criterion, and scalar-weight analyses may miss higher-order interactions or redundant pathways. The theoretical analysis should likewise be read as a local support-recovery argument for simplified linear surrogates. A stronger account would need to connect feature coherence, exact sparsity, representation geometry, and training-time dynamics in a single framework, and to specify when sparsity promotes separability versus when it destroys useful shared structure. Extending the framework to larger models should also test stability across extraction settings and datasets, compare weight-, path-, and circuit-level interventions, and evaluate targeted functional editing beyond masking.

\section{Methods}

\subsection{Model Architectures}

We conduct experiments using two representative neural architectures: a multilayer perceptron (MLP) and a lightweight transformer model (\textit{TinyTransformer}).

\subsubsection{Multitask MLP}

The multitask MLP consists of three fully connected hidden layers, each containing 300 ReLU units (Fig.~\ref{fig:main1}a). Each training sample comprises a shared input pair and labels for all selected tasks, and the output layer concatenates $R$ rule-specific predictions corresponding to $R$ distinct tasks. In the default setting, $R=5$, and the tasks consist of related arithmetic or logical transformations applied to the same input pair. All tasks are introduced simultaneously and optimized jointly from random initialization, without curriculum learning, sequential task presentation, or task-specific pretraining. Hidden activations are used for representation analyses, whereas functional-token extraction is performed on scalar weight coordinates in the trained parameter matrices. Sparsity is imposed using $L_\gamma$ regularization with $\gamma\in\{0.5,1,2\}$, with $\gamma=1$ used as the default unless otherwise stated. The task loss is averaged over the equally sized rule outputs, and the regularization penalty is the sum of the mean $(|w|+\varepsilon)^\gamma$ within each trainable non-bias weight matrix, with $\varepsilon=10^{-8}$. Additional architectural and training details, including the exact objective, are provided in Supplemental Information, Section~\ref{appendix:architecture}.

\subsubsection{TinyTransformer}

To test whether functional modularity generalizes beyond MLPs, we also consider a lightweight transformer encoder (Fig.~\ref{fig:main3}e). The model embeds the input into a compact sequence representation, applies a lightweight transformer encoder, and maps the pooled representation to the same multitask output format used by the MLP experiments. The TinyTransformer control is trained on Binary Sequence Calculation. As in the MLP experiments, all tasks are trained jointly from random initialization, without curriculum learning or sequential task presentation. Functional-token extraction is performed on scalar weight coordinates. The TinyTransformer analysis includes 23 independent runs trained with batch size 128. Full implementation details are provided in Supplemental Information, Section~\ref{appendix:architecture}.

\subsection{Task Design and Random Multitask Training}
\label{method:task_design}

We use several rule-based multitask families to test whether functional tokens emerge across different input formats and rule semantics (Supplemental Information, Section~\ref{appendix:tasks}). In all cases, a single shared neural network receives one input sample and predicts the concatenated outputs of the selected rules. Unless otherwise stated, experiments use the first five rules of the selected task family as the default multitask setting ($R=5$).

For Binary Sequence Calculation tasks, each input consists of two binary vectors of length $P=7$. The complete task family contains six arithmetic rules. Rules~1--4 are primitive transformations--addition, multiplication, $a^2$, and $b^2$--whereas rules~5 and~6 are composite transformations, $a(a+b)$ and $b(a+b)$, which reuse primitive arithmetic structure. The default setting jointly trains rules~1--5, placing four primitive tasks and one composite task in the same shared network.

Sparse Pattern Logic tasks use the same paired binary-input format but replace arithmetic with elementwise logical operations. The full rule family contains AND, OR, XOR, NOT of the first input, identity of the first input, and identity of the second input; the default setting again uses the first five rules. MNIST Digit-Pair Arithmetic tasks mirror the Binary Sequence Calculation rule structure but use digit-derived inputs, testing whether the same extraction framework remains informative beyond direct synthetic binary variables. The TinyTransformer control uses the Binary Sequence Calculation setting with the same five default rules.

For each independent trial, examples are randomly sampled and split into training and validation sets. Labels for all selected rules are generated on the same samples, and all task losses are optimized jointly from random initialization. Additional implementation details are provided in Supplemental Information, Section~\ref{appendix:architecture}.

\subsection{Profiling Functional Tokens}

\subsubsection{Ablation Profiling}
For each trained network and rule $r$, we quantify the functional contribution of a scalar weight token $q$ by zeroing that weight while keeping all other parameters fixed. Let $f_r(x)$ denote the baseline output block for rule $r$, and let $f_r^{(-q)}(x)$ denote the output after ablating scalar weight $q$. The ablation score is
\begin{equation}
S^{\mathrm{abl}}_{r,q}
=\mathbb{E}_{x\sim\mathcal{D}_r}
\left\|f_r(x)-f_r^{(-q)}(x)\right\|_2^2 .
\end{equation}
This score measures the output perturbation induced by removing one scalar weight coordinate from the trained model.

\subsubsection{Taylor Profiling}
As a complementary first-order approximation, we estimate the squared output sensitivity of each scalar weight:
\begin{equation}
S^{\mathrm{tay}}_{r,q}=\theta_q^2\,\mathbb{E}_x\left\|\frac{\partial f_r(x;\theta)}{\partial\theta_q}\right\|_2^2.
\label{eq:taylor_output}
\end{equation}
This approximates the squared first-order output perturbation associated with removing a scalar weight. For each batch $B$, rule $r$ and probe $k=1,\ldots,K$, independent Rademacher signs $z_{i,d}^{B,k,r}\in\{-1,+1\}$ are drawn for each sample $i$ and output coordinate $d$. The estimator is
\begin{equation}
g_{B,k,r,q}=\frac{\partial}{\partial\theta_q}\sum_{i\in B}\sum_{d=1}^{P}z_{i,d}^{B,k,r}f_{r,d}(x_i;\theta),\qquad
\widehat S_{r,q}=\frac{1}{NK}\sum_B\sum_{k=1}^{K}(\theta_q g_{B,k,r,q})^2.
\label{eq:taylor_probe}
\end{equation}
Here $N$ is the total number of profiled samples. The expectation over the probes equals $\theta_q^2N^{-1}\sum_i\|\partial f_r(x_i)/\partial\theta_q\|_2^2$. The Taylor runs in Figs.~\ref{fig:main3}c,d,f and~\ref{fig:main5}a used $K=4$.

\subsection{Functional Token Extraction}

A weight token is a scalar coordinate in a trainable non-bias weight matrix. A rule-specific functional token is the set of weight tokens selected by an intervention-defined important mask. For each rule $r$ and weight token $q$, let $S_{r,q}\ge 0$ denote the corresponding importance score from ablation or Taylor profiling. To account for differences in output scale across rules, we also compute a normalized score profile
\begin{equation}
G_r=\mathbb{E}_{x\sim\mathcal{D}_r}\|f_r(x)\|_2^2+\epsilon_{\mathrm{out}},
\qquad
Q_{r,q}=\frac{S_{r,q}}{G_r},
\end{equation}
where $\epsilon_{\mathrm{out}}$ stabilizes output-energy normalization; it is distinct from the score floor $\epsilon_S$.

The extraction quantile is fixed at $\tau=0.75$. For each trained network $n$, profiled weight matrix $b$ and rule $r$, the numerical threshold is
\begin{equation}
T_{n,b,r}=\operatorname{Quantile}_{0.75}\{S_{n,b,r,q}:q\in\mathcal C_{n,b},\ S_{n,b,r,q}\text{ finite}\},
\end{equation}
where $\mathcal C_{n,b}$ is the set of profiled candidate coordinates. The important mask is
\begin{equation}
I_{n,b,r,q}=\mathbf{1}\{S_{n,b,r,q}\text{ finite}\}\mathbf{1}\{S_{n,b,r,q}>\epsilon_S\}\mathbf{1}\{S_{n,b,r,q}\ge T_{n,b,r}\}.
\label{eq:important_mask}
\end{equation}
The quantile is computed before score-floor filtering, so finite zero scores enter its calculation. The numerical threshold is recomputed for every network, matrix and rule, and at every profiled training snapshot. For a rule set $C$, masks are combined by union,
\begin{equation}
I_{C,b,q}=\max_{r\in C}I_{b,r,q}.
\end{equation}
We construct a mask separately for each profiled weight matrix and apply all masks simultaneously, including the output weights. Biases and other parameters outside the masking set remain unchanged. The retained-weight ratio is
\begin{equation}
\rho_{\mathrm{mask}}(C)=\frac{\sum_b\sum_{q\in W_b}I_{C,b,q}}{\sum_b|W_b|}.
\label{eq:retained_mask_ratio}
\end{equation}
Its numerator counts mask-selected positions; its denominator counts all scalar coordinates in the analyzed weight matrices, including zero-valued entries and excluding biases.

We evaluate rule-wise \emph{relative accuracy} as
\begin{equation}
A(C,r) = \frac{A'(C,r)}{A_0(r)},
\label{eq:eq3}
\end{equation}
where $A'(C,r)$ is the accuracy of the retained subnetwork on rule $r$, and $A_0(r)$ is the accuracy of the full network on the same rule. Unless otherwise stated, rule accuracy is the mean agreement over samples and output bits after binarizing predictions at $>0.5$.

For descriptive assignment plots, a selected weight token can also be given an exact rule-set signature,
\begin{equation}
\sigma(q)=\{r:I_{r,q}=1\}.
\end{equation}
These signature counts visualize which rule combinations share selected weights.

The importance-weighted normalized inverse participation (NIP) statistic used in Fig.~\ref{fig:main4}c,d and the training trajectories in Fig.~\ref{fig:main5}a uses the masked score profile directly. Given a rule set $\mathcal{R}$ and a block-specific important mask $I_{r,q}$, define
\begin{equation}
V_{r,q}=I_{r,q}\max(Q_{r,q},0),
\qquad a_q=\sum_{r\in\mathcal{R}}V_{r,q}.
\end{equation}
Tokens are active when $a_q>\epsilon_M$, with $\epsilon_M=10^{-12}$, a numerical mass floor distinct from the raw-score floor $\epsilon_S$. For active tokens, let $p_{r,q}=V_{r,q}/a_q$. We define the normalized concentration
\begin{equation}
\phi_q=\sum_{r\in\mathcal{R}}p_{r,q}^2,
\qquad
z_q=\frac{\phi_q-1/|\mathcal{R}|}{1-1/|\mathcal{R}|},
\end{equation}
and the block modularity score
\begin{equation}
M=\frac{\sum_{q:a_q>\epsilon_M}a_q z_q}{\sum_{q:a_q>\epsilon_M}a_q}.
\label{eq:modularity}
\end{equation}
Thus $M\in[0,1]$ measures the degree to which important weighted score profiles are concentrated on individual rules, with larger values indicating stronger functional specialization under a fixed extraction criterion. Masks are recomputed separately for each network, weight matrix, rule and snapshot. $M$ is defined when the active importance mass is nonzero.

Additional implementation details are provided in Supplementary Note~\ref{appendix:Taylor}.

\subsubsection*{Supplemental Information}

Supplemental theoretical analysis, model and training details, task definitions, and supporting analyses are appended after the References.

\subsubsection*{Data and Code Availability}

All synthetic datasets are generated by the task procedures described in Methods and Supplemental Information. MNIST-based experiments use the public MNIST dataset. Training, analysis, and figure-generation code are maintained in the accompanying project repository.

\subsubsection*{Acknowledgements}

This work was supported by the Beijing Natural Science Foundation (Z260005 to L. Q.) and the Fundamental and Interdisciplinary Disciplines Breakthrough Plan of the Ministry of Education of China (JYB2025XDXM502 to Q. O.).

\subsubsection*{Author contributions}

L. Q. and Q. O. conceived the study and designed the overall research framework. B. S. developed the theoretical methodology, designed and carried out the computational experiments and conducted data analysis. L. Q. supervised the research. B. S. and L. Q. wrote the manuscript with input from all authors. All authors discussed the results, reviewed the manuscript, and approved the final version.

\subsubsection*{Competing interests}

The authors declare no competing interests.

\bibliographystyle{unsrtnat}

\bibliography{ref}

\clearpage
\renewcommand{\theHsection}{S.\arabic{section}}
\renewcommand{\theHequation}{S.\arabic{equation}}
\renewcommand{\theHfigure}{S.\arabic{figure}}
\renewcommand{\theHtable}{S.\arabic{table}}
\title{Supplemental Information for\\Sparsity-driven Emergence of Functional Modularity in Multitask Neural Networks}

\presubmaketitle

\setcounter{section}{0}
\setcounter{equation}{0}
\setcounter{figure}{0}
\setcounter{table}{0}
\renewcommand{\thesection}{S\arabic{section}}
\renewcommand{\thesubsection}{\thesection.\arabic{subsection}}
\renewcommand{\thesubsubsection}{\thesubsection.\arabic{subsubsection}}
\renewcommand{\theequation}{S.\arabic{equation}}
\renewcommand{\thefigure}{S\arabic{figure}}
\renewcommand{\thetable}{S\arabic{table}}

\section{A Local Support-Recovery Argument for \texorpdfstring{\(L_\gamma\)}{L-gamma}-Induced Subnetworks}
\label{appendix:lgamma_subnetwork_recovery}
\label{appendix:proofLasso}

This supplemental section gives a conditional proof explaining why sparsity-inducing \(L_\gamma\) regularization can make task-selective functional-token subnetworks separable by ablation or Taylor profiling. The argument starts from the linear multitask model
\begin{equation}
Y=XB_0+E,
\label{eq:appendix_linear_model}
\end{equation}
and then transfers the conclusion to a trained neural network through an explicit local-linear surrogate assumption. It should be read as a support-recovery theorem for the surrogate coordinates, not as a global theorem for nonlinear training dynamics.

\subsection{Linear surrogate and notation}

Let \(n\) be the number of samples, \(p\) the number of candidate elementary coordinates, and \(R\) the number of tasks. In Eq.~\eqref{eq:appendix_linear_model},
\begin{equation}
X\in\mathbb R^{n\times p},\qquad
Y=[Y_1,\ldots,Y_R]\in\mathbb R^{n\times R},\qquad
B_0=[\beta_{0,1},\ldots,\beta_{0,R}]\in\mathbb R^{p\times R},
\end{equation}
where \(Y_r\in\mathbb R^n\) is the target vector for task \(r\), \(\beta_{0,r}\in\mathbb R^p\) is the true task-specific coefficient vector, and \(E=[\varepsilon_1,\ldots,\varepsilon_R]\) is the noise matrix. For each task,
\begin{equation}
Y_r=X\beta_{0,r}+\varepsilon_r .
\end{equation}
We write
\begin{equation}
C=\frac{1}{n}X^\top X,
\qquad
\mathcal S_r=\{j:\beta_{0,jr}\ne0\},
\qquad
\mathcal S_r^c=\{1,\ldots,p\}\setminus\mathcal S_r,
\qquad
s_r=|\mathcal S_r|.
\end{equation}
Here \(\mathcal S_r\) is the true support of task \(r\). To avoid confusing support sets with importance scores, this section uses calligraphic \(\mathcal S\) for supports and reserves symbols such as \(S^{\mathrm{abl}}_{r,q}\) and \(S^{\mathrm{tay}}_{r,q}\) for ablation and Taylor scores.

We estimate \(B_0\) by
\begin{equation}
\widehat B_\gamma\in
\arg\min_B
Q_\gamma(B),
\qquad
Q_\gamma(B)=
\frac{1}{2n}\|Y-XB\|_F^2+
\lambda\sum_{j=1}^{p}\sum_{r=1}^{R}|B_{jr}|^\gamma,
\qquad 0<\gamma\le1 .
\label{eq:lgamma_multitask_objective}
\end{equation}
The objective is separable across task columns:
\begin{equation}
Q_\gamma(B)=\sum_{r=1}^{R}q_{\gamma,r}(\beta_r),
\qquad
q_{\gamma,r}(\beta)=
\frac{1}{2n}\|Y_r-X\beta\|_2^2+
\lambda\sum_{j=1}^{p}|\beta_j|^\gamma .
\end{equation}
It is therefore enough to prove the support claim for one fixed task. In the proof below we suppress the task index \(r\), writing \(Y,\beta_0,\varepsilon,\mathcal S\) for the corresponding single-task quantities.

\subsection{Assumptions}

The following assumptions are standard support-recovery assumptions, with an additional local-curvature condition for the nonconvex case \(0<\gamma<1\).

\paragraph{A1. Sparse and normalized design.}
The support size satisfies \(s=|\mathcal S|\ll p\). The columns of \(X\) are normalized so that \(C_{jj}=1\), and \(\|C\|_{op}\le M_C\) for a finite constant \(M_C\). Here \(\|\cdot\|_{op}\) denotes the spectral norm.

\paragraph{A2. Restricted identifiability.}
There are constants \(m>0\) and \(M_S<\infty\) such that
\begin{equation}
\lambda_{\min}(C_{\mathcal S\mathcal S})\ge m,
\qquad
\|C_{\mathcal S\mathcal S}^{-1}\|_\infty\le M_S .
\label{eq:restricted_identifiability_lgamma}
\end{equation}
For index sets \(A,B\), \(C_{AB}\) denotes the submatrix of \(C\) with rows in \(A\) and columns in \(B\). The matrix \(\infty\)-norm is the maximum absolute row sum.

\paragraph{A3. Noise correlation.}
The empirical noise-design correlation is uniformly bounded:
\begin{equation}
\left\|\frac{1}{n}X^\top\varepsilon\right\|_\infty\le\tau_n .
\label{eq:noise_correlation_lgamma}
\end{equation}
For normalized designs and sub-Gaussian noise, \(\tau_n\) is typically of order \(\sqrt{\log p/n}\) for one task, or \(\sqrt{\log(pR)/n}\) uniformly across \(R\) tasks.

\paragraph{A4. Beta-min signal.}
The smallest active coefficient is bounded below:
\begin{equation}
b_{\min}=\min_{j\in\mathcal S}|\beta_{0,j}|>0 .
\end{equation}
The tuning parameter \(\lambda\) is small enough that active coefficients do not cross zero. A convenient sufficient condition is
\begin{equation}
M_S\{\tau_n+\lambda\gamma(b_{\min}/2)^{\gamma-1}\}<b_{\min}/2 .
\label{eq:beta_min_lgamma_subnetwork}
\end{equation}
For \(\gamma=1\), this is \(M_S(\tau_n+\lambda)<b_{\min}/2\).

\paragraph{A5. Off-support incoherence for \(\gamma=1\).}
For the Lasso endpoint, assume the strict irrepresentable condition
\begin{equation}
\left\|
C_{\mathcal S^c\mathcal S}C_{\mathcal S\mathcal S}^{-1}
\operatorname{sign}(\beta_{0,\mathcal S})
\right\|_\infty
\le1-\eta,
\qquad 0<\eta\le1,
\label{eq:irrepresentable_lgamma_subnetwork}
\end{equation}
and assume the off-support noise is dominated by the Lasso penalty:
\begin{equation}
\left(1+\left\|C_{\mathcal S^c\mathcal S}C_{\mathcal S\mathcal S}^{-1}\right\|_\infty\right)\tau_n
\le \eta\lambda .
\label{eq:lasso_noise_dominance_lgamma}
\end{equation}
These two inequalities are used only for the convex \(\gamma=1\) support-recovery proof.

\paragraph{A6. Active-set curvature for \(0<\gamma<1\).}
For \(0<\gamma<1\), \(|b|^\gamma\) is concave away from zero, so its second derivative is negative on active coordinates. We assume the restricted quadratic loss curvature dominates that negative curvature in the sign-preserving neighborhood \(|b_j|\ge b_{\min}/2\): there is \(\mu>0\) such that
\begin{equation}
\lambda_{\min}(C_{\mathcal S\mathcal S})
-
\lambda\gamma(1-\gamma)(b_{\min}/2)^{\gamma-2}
\ge \mu .
\label{eq:active_curvature_lgamma_subnetwork}
\end{equation}
In addition, the active gradient at the oracle coefficient is small enough that
\begin{equation}
\sqrt{s}\left\{\tau_n+\lambda\gamma(b_{\min}/2)^{\gamma-1}\right\}<\mu b_{\min}/4 .
\label{eq:nonconvex_boundary_lgamma}
\end{equation}
This condition ensures that the restricted nonconvex problem has an interior minimizer in the sign-preserving neighborhood.

\subsection{Support recovery for \texorpdfstring{\(0<\gamma\le1\)}{0 < gamma <= 1}}

\paragraph{Proposition S1.}
Under A1--A6, the following statements hold for each task \(r\).

For \(\gamma=1\), the convex Lasso problem has a global minimizer \(\widehat\beta_{1,r}\) satisfying
\begin{equation}
\operatorname{supp}(\widehat\beta_{1,r})=\mathcal S_r,
\qquad
\operatorname{sign}(\widehat\beta_{1,\mathcal S_r,r})=
\operatorname{sign}(\beta_{0,\mathcal S_r,r}).
\end{equation}

For \(0<\gamma<1\), the nonconvex objective has a strict local minimizer \(\widehat\beta_{\gamma,r}^{\mathrm{loc}}\) satisfying
\begin{equation}
\operatorname{supp}(\widehat\beta_{\gamma,r}^{\mathrm{loc}})=\mathcal S_r,
\qquad
\operatorname{sign}(\widehat\beta_{\gamma,\mathcal S_r,r}^{\mathrm{loc}})=
\operatorname{sign}(\beta_{0,\mathcal S_r,r}).
\end{equation}
The second statement is local: it does not claim that all global minimizers have this support, nor that gradient descent from arbitrary initialization must reach this local minimizer.

\paragraph{Proof.}
Fix a task and omit the subscript \(r\).

\emph{Case 1: \(\gamma=1\).}
The objective is convex. We use a primal-dual witness argument. The phrase Karush-Kuhn-Tucker (KKT) conditions refers here to the necessary and sufficient first-order optimality conditions for a convex optimization problem with a nonsmooth penalty, expressed through the subgradient of the \(\ell_1\) norm.

First solve the restricted Lasso problem on the true support:
\begin{equation}
\widehat\beta_{\mathcal S}
\in
\arg\min_{b\in\mathbb R^s}
\left\{
\frac{1}{2n}\|Y-X_{\mathcal S}b\|_2^2+
\lambda\|b\|_1
\right\},
\qquad
\widehat\beta_{\mathcal S^c}=0 .
\end{equation}
On active coordinates, the KKT equation is
\begin{equation}
C_{\mathcal S\mathcal S}(\widehat\beta_{\mathcal S}-\beta_{0,\mathcal S})
-
\frac{1}{n}X_{\mathcal S}^\top\varepsilon
+
\lambda z_{\mathcal S}=0,
\label{eq:kkt_active_lasso_lgamma}
\end{equation}
where \(z_{\mathcal S}\in\partial\|\widehat\beta_{\mathcal S}\|_1\). If \(\widehat\beta_{\mathcal S}\) has the correct sign, then \(z_{\mathcal S}=\operatorname{sign}(\beta_{0,\mathcal S})\). Solving Eq.~\eqref{eq:kkt_active_lasso_lgamma} gives
\begin{equation}
\widehat\beta_{\mathcal S}-\beta_{0,\mathcal S}
=
C_{\mathcal S\mathcal S}^{-1}
\left(
\frac{1}{n}X_{\mathcal S}^\top\varepsilon-
\lambda z_{\mathcal S}
\right).
\end{equation}
By A2--A4 with \(\gamma=1\),
\begin{equation}
\|\widehat\beta_{\mathcal S}-\beta_{0,\mathcal S}\|_\infty
\le M_S(\tau_n+\lambda)<b_{\min}/2 .
\end{equation}
Thus every active coefficient remains nonzero and has the same sign as \(\beta_{0,\mathcal S}\). This validates the chosen active subgradient.

It remains to verify the inactive KKT condition. For \(j\in\mathcal S^c\), the subgradient condition for \(\widehat\beta_j=0\) is
\begin{equation}
\left\|
C_{\mathcal S^c\mathcal S}(\widehat\beta_{\mathcal S}-\beta_{0,\mathcal S})
-
\frac{1}{n}X_{\mathcal S^c}^\top\varepsilon
\right\|_\infty
\le\lambda .
\label{eq:kkt_inactive_lasso_lgamma}
\end{equation}
Substituting the active solution with \(z_{\mathcal S}=\operatorname{sign}(\beta_{0,\mathcal S})\), the left side of Eq.~\eqref{eq:kkt_inactive_lasso_lgamma} is at most
\begin{equation}
\lambda
\left\|
C_{\mathcal S^c\mathcal S}C_{\mathcal S\mathcal S}^{-1}
\operatorname{sign}(\beta_{0,\mathcal S})
\right\|_\infty
+
\left(1+
\left\|C_{\mathcal S^c\mathcal S}C_{\mathcal S\mathcal S}^{-1}\right\|_\infty
\right)\tau_n .
\end{equation}
By A5 this is bounded by \(\lambda(1-\eta)+\eta\lambda=\lambda\). Therefore the full vector \((\widehat\beta_{\mathcal S},0)\) satisfies the KKT conditions. Since the \(\gamma=1\) objective is convex, these conditions certify a global minimizer with support \(\mathcal S\) and correct signs.

\emph{Case 2: \(0<\gamma<1\).}
The penalty is nonconvex and non-Lipschitz at zero, so KKT conditions at zero are not the right tool. We instead prove the existence of an oracle-supported strict local minimizer.

Consider the restricted objective on the true support:
\begin{equation}
q_{\gamma,\mathcal S}(b)=
\frac{1}{2n}\|Y-X_{\mathcal S}b\|_2^2+
\lambda\sum_{j\in\mathcal S}|b_j|^\gamma .
\end{equation}
Define the sign-preserving neighborhood
\begin{equation}
\mathcal N_{\mathcal S}=\{b:\|b-\beta_{0,\mathcal S}\|_\infty\le b_{\min}/2\}.
\end{equation}
Inside this neighborhood all active coordinates satisfy \(|b_j|\ge b_{\min}/2\). For any point in \(\mathcal N_{\mathcal S}\), the Hessian of the restricted objective is
\begin{equation}
\nabla^2 q_{\gamma,\mathcal S}(b)
=
C_{\mathcal S\mathcal S}
-
\lambda\gamma(1-\gamma)
\operatorname{diag}(|b_j|^{\gamma-2}:j\in\mathcal S),
\end{equation}
which is bounded below by \(\mu I\) by A6. Hence \(q_{\gamma,\mathcal S}\) is \(\mu\)-strongly convex on \(\mathcal N_{\mathcal S}\), even though the full objective is nonconvex outside this neighborhood.

At the oracle coefficient,
\begin{equation}
g_0=
\nabla q_{\gamma,\mathcal S}(\beta_{0,\mathcal S})
=
-
\frac{1}{n}X_{\mathcal S}^\top\varepsilon
+
\lambda\gamma|\beta_{0,\mathcal S}|^{\gamma-1}
\operatorname{sign}(\beta_{0,\mathcal S}).
\end{equation}
By A3 and A6,
\begin{equation}
\|g_0\|_2
\le
\sqrt{s}\{\tau_n+\lambda\gamma(b_{\min}/2)^{\gamma-1}\}
<\mu b_{\min}/4 .
\end{equation}
For any boundary point \(b=\beta_{0,\mathcal S}+u\) of \(\mathcal N_{\mathcal S}\), \(\|u\|_2\ge b_{\min}/2\). Strong convexity gives
\begin{equation}
q_{\gamma,\mathcal S}(\beta_{0,\mathcal S}+u)-q_{\gamma,\mathcal S}(\beta_{0,\mathcal S})
\ge
u^\top g_0+\frac{\mu}{2}\|u\|_2^2
\ge
-\|u\|_2\|g_0\|_2+\frac{\mu}{2}\|u\|_2^2>0 .
\end{equation}
Therefore the minimizer of \(q_{\gamma,\mathcal S}\) over the compact set \(\mathcal N_{\mathcal S}\) lies in its interior. Denote it by \(\widehat\beta_{\mathcal S}\). It satisfies the active first-order equation
\begin{equation}
C_{\mathcal S\mathcal S}(\widehat\beta_{\mathcal S}-\beta_{0,\mathcal S})
-
\frac{1}{n}X_{\mathcal S}^\top\varepsilon
+
\lambda\gamma|\widehat\beta_{\mathcal S}|^{\gamma-1}
\operatorname{sign}(\widehat\beta_{\mathcal S})=0 .
\label{eq:active_nonconvex_lgamma}
\end{equation}
Because \(\widehat\beta_{\mathcal S}\in\mathcal N_{\mathcal S}\), all active signs are preserved.

We now show that \((\widehat\beta_{\mathcal S},0)\) is a strict local minimizer in the full \(p\)-dimensional space. Perturb the active and inactive coordinates by \(u\in\mathbb R^s\) and \(v\in\mathbb R^{p-s}\), respectively. A Taylor expansion around \((\widehat\beta_{\mathcal S},0)\), using Eq.~\eqref{eq:active_nonconvex_lgamma} for the active coordinates, yields
\begin{align}
\Delta q(u,v)
&=
q_\gamma(\widehat\beta_{\mathcal S}+u,v)-q_\gamma(\widehat\beta_{\mathcal S},0) \\
&\ge
\frac{\mu}{2}\|u\|_2^2
+
 u^\top C_{\mathcal S\mathcal S^c}v
+
 a^\top v
-
\frac{M_C}{2}\|v\|_2^2
+
\lambda\sum_{j\in\mathcal S^c}|v_j|^\gamma,
\label{eq:delta_full_nonconvex_lgamma}
\end{align}
where
\begin{equation}
a=
C_{\mathcal S^c\mathcal S}(\widehat\beta_{\mathcal S}-\beta_{0,\mathcal S})
-
\frac{1}{n}X_{\mathcal S^c}^\top\varepsilon
\end{equation}
is finite by A1--A4. For \(0<\gamma<1\), the term \(\sum_j|v_j|^\gamma\) dominates both \(\|v\|_1\) and \(\|v\|_2^2\) as \(\|v\|_\infty\to0\): for sufficiently small \(\|v\|_\infty\),
\begin{equation}
\|v\|_1\le \|v\|_\infty^{1-\gamma}\sum_j|v_j|^\gamma,
\qquad
\|v\|_2^2\le \|v\|_\infty^{2-\gamma}\sum_j|v_j|^\gamma .
\end{equation}
The cross term satisfies, for any \(\alpha>0\),
\begin{equation}
|u^\top C_{\mathcal S\mathcal S^c}v|
\le
\alpha\|u\|_2^2+
\frac{\|C_{\mathcal S\mathcal S^c}\|_{op}^2}{4\alpha}\|v\|_2^2 .
\end{equation}
Choose \(\alpha<\mu/4\), then choose the neighborhood in \(v\) small enough that the inactive singular penalty dominates the finite linear term \(a^\top v\), the quadratic term in \(v\), and the remaining cross-term bound. Eq.~\eqref{eq:delta_full_nonconvex_lgamma} is then strictly positive for every nonzero sufficiently small perturbation \((u,v)\). Thus \((\widehat\beta_{\mathcal S},0)\) is a strict local minimizer with support \(\mathcal S\). This proves the nonconvex local statement.

Repeating the single-task argument for each task column proves the multitask statement, because Eq.~\eqref{eq:lgamma_multitask_objective} separates across tasks.

\subsection{From support recovery to score-separable functional tokens}

Support recovery says that the effective coordinates of task \(r\) form a set \(\mathcal S_r\). The experiments, however, do not observe \(\beta_{0,r}\) directly. They estimate functional supports through ablation and Taylor scores. This subsection proves the link between the true support \(\mathcal S_r\), the ablation score \(S^{\mathrm{abl}}_{r,q}\), and the Taylor score \(S^{\mathrm{tay}}_{r,q}\).

Let \(f_\theta\) be a trained network and fix a local coordinate block \(\ell\). Let \(z_q(x)\) denote the scalar local coordinate associated with weight token \(q\), for example the coordinate appearing in a parameter-Jacobian linearization around the trained parameters. Let \(f_r(x)\in\mathbb R^d\) be the output segment for task \(r\), where \(d=P\) in the experiments. Assume the following local-linear surrogate:
\begin{equation}
f_r(x)=\sum_{q=1}^{d_\ell}z_q(x)b_{qr}+\xi_r(x),
\qquad
b_{qr}\in\mathbb R^d,
\label{eq:local_linear_hidden_lgamma}
\end{equation}
where \(\xi_r(x)\) is a remainder term. The task support in this layer is
\begin{equation}
\mathcal S_{\ell,r}=\{q:\|b_{qr}\|_2>0\}.
\end{equation}
Define the ideal coefficient-energy score
\begin{equation}
G_{r,q}=\mathbb E\left[z_q(x)^2\right]\|b_{qr}\|_2^2 .
\label{eq:ideal_energy_score_lgamma}
\end{equation}
Assume an active margin: there exists \(\kappa>0\) such that
\begin{equation}
q\in\mathcal S_{\ell,r}\quad\Rightarrow\quad G_{r,q}\ge\kappa,
\qquad
q\notin\mathcal S_{\ell,r}\quad\Rightarrow\quad G_{r,q}=0 .
\label{eq:active_score_margin_lgamma}
\end{equation}
This margin excludes unidentifiable active coordinates with zero coordinate second moment or vanishingly small effective coefficients.

\paragraph{Ablation score.}
Let \(f_r^{(-q)}(x)\) be the output after removing token \(q\). The population ablation score is
\begin{equation}
S^{\mathrm{abl}}_{r,q}=
\mathbb E\left\|f_r(x)-f_r^{(-q)}(x)\right\|_2^2 .
\label{eq:population_ablation_score_lgamma}
\end{equation}
If removing \(q\) changes only the corresponding local term up to a remainder
\begin{equation}
e^{\mathrm{abl}}_{r,q}(x)=
f_r(x)-f_r^{(-q)}(x)-z_q(x)b_{qr},
\qquad
\mathbb E\|e^{\mathrm{abl}}_{r,q}(x)\|_2^2\le \epsilon_{\mathrm{abl}}^2,
\end{equation}
and if \(G_{r,q}\le M_G\) uniformly, then
\begin{align}
\left|S^{\mathrm{abl}}_{r,q}-G_{r,q}\right|
&=
\left|\mathbb E\|z_qb_{qr}+e^{\mathrm{abl}}_{r,q}\|_2^2-
\mathbb E\|z_qb_{qr}\|_2^2\right| \\
&\le
2\sqrt{M_G}\epsilon_{\mathrm{abl}}+
\epsilon_{\mathrm{abl}}^2
=:\delta_{\mathrm{abl}} .
\label{eq:ablation_energy_bound_lgamma}
\end{align}
Thus ablation score is a controlled perturbation of the ideal support energy \(G_{r,q}\).

\paragraph{Taylor score.}
Introduce a multiplicative gate \(m_q\) on token \(q\), so that the local surrogate becomes
\begin{equation}
f_r(x;m)=\sum_{k=1}^{d_\ell}m_kz_k(x)b_{kr}+\xi_r(x;m),
\qquad m_k=1\ \text{at the trained network.}
\end{equation}
The population first-order Taylor score is
\begin{equation}
S^{\mathrm{tay}}_{r,q}=
\mathbb E\left\|
\left.
\frac{\partial f_r(x;m)}{\partial m_q}
\right|_{m=\mathbf 1}
\right\|_2^2 .
\label{eq:population_taylor_score_lgamma}
\end{equation}
Because
\begin{equation}
\left.
\frac{\partial f_r(x;m)}{\partial m_q}
\right|_{m=\mathbf 1}
=
z_q(x)b_{qr}+e^{\mathrm{tay}}_{r,q}(x),
\qquad
e^{\mathrm{tay}}_{r,q}(x)=
\left.
\frac{\partial \xi_r(x;m)}{\partial m_q}
\right|_{m=\mathbf 1},
\end{equation}
the same calculation gives, if \(\mathbb E\|e^{\mathrm{tay}}_{r,q}(x)\|_2^2\le\epsilon_{\mathrm{tay}}^2\),
\begin{equation}
\left|S^{\mathrm{tay}}_{r,q}-G_{r,q}\right|
\le
2\sqrt{M_G}\epsilon_{\mathrm{tay}}+
\epsilon_{\mathrm{tay}}^2
=:\delta_{\mathrm{tay}} .
\label{eq:taylor_energy_bound_lgamma}
\end{equation}

\paragraph{Score-support relation.}
Define score-threshold support estimates
\begin{equation}
\widehat{\mathcal S}^{\mathrm{abl}}_{\ell,r}(t)=
\{q:S^{\mathrm{abl}}_{r,q}>t\},
\qquad
\widehat{\mathcal S}^{\mathrm{tay}}_{\ell,r}(t)=
\{q:S^{\mathrm{tay}}_{r,q}>t\}.
\end{equation}
Let \(\delta=\max\{\delta_{\mathrm{abl}},\delta_{\mathrm{tay}}\}\). If \(\delta<\kappa/2\), then for every threshold
\begin{equation}
t\in(\delta,\kappa-\delta)
\end{equation}
we have
\begin{equation}
\widehat{\mathcal S}^{\mathrm{abl}}_{\ell,r}(t)
=
\widehat{\mathcal S}^{\mathrm{tay}}_{\ell,r}(t)
=
\mathcal S_{\ell,r}.
\label{eq:score_support_equivalence_lgamma}
\end{equation}
Indeed, if \(q\notin\mathcal S_{\ell,r}\), then \(G_{r,q}=0\), so both scores are at most \(\delta<t\). If \(q\in\mathcal S_{\ell,r}\), then \(G_{r,q}\ge\kappa\), so both scores are at least \(\kappa-\delta>t\). Moreover,
\begin{equation}
\left|S^{\mathrm{abl}}_{r,q}-S^{\mathrm{tay}}_{r,q}\right|
\le
\delta_{\mathrm{abl}}+\delta_{\mathrm{tay}},
\end{equation}
so ablation and Taylor are close because both estimate the same ideal energy \(G_{r,q}\), not because they are algebraically identical interventions.

The empirical scores have finite-sample error and, for Taylor profiling, random-probe error. If uniform bounds of sizes $c_N$ and $c_K$ hold for these two errors, the same conclusion holds after replacing $\delta$ by $\delta+c_N+c_K$. Four probes do not automatically guarantee the required bound; no numerical probe-error guarantee is asserted here. This is the formal relation between the recovered task support \(\mathcal S_{\ell,r}\), the ablation score \(S^{\mathrm{abl}}_{r,q}\), and the Taylor score \(S^{\mathrm{tay}}_{r,q}\).

\subsection{From linear supports to functional-token subnetworks}

The support-recovery result applies directly to coordinates on which the regularized objective acts, such as readout coefficients, explicit gates, path coefficients, or parameter-Jacobian coordinates. For scalar weight coordinates in a nonlinear network, raw sparsity must still be interpreted through a local surrogate. The bridge from Eq.~\eqref{eq:appendix_linear_model} to a neural network therefore requires a compatibility assumption: in the chosen local coordinates, the trained network is well approximated by Eq.~\eqref{eq:local_linear_hidden_lgamma}, and the effect of \(L_\gamma\) regularization is reflected in sparse, stable effective coefficients \(b_{qr}\) with a nonzero score margin \(\kappa\).

The support-recovery theorem and its singularity argument for $0<\gamma<1$ concern the idealized, unshifted penalty $|\beta|^\gamma$. The neural-network implementation instead uses $(|\theta|+10^{-8})^\gamma$ with per-matrix normalization (Section~\ref{appendix:architecture}). For this shifted penalty, the one-sided derivative has finite magnitude at zero; the unshifted singularity argument therefore does not directly prove local stability for the implemented objective. The theoretical connection remains conditional on the local-surrogate and compatibility assumptions above.

The support-recovery statement applies only when the realized score threshold, including the score floor, lies inside the assumed separation gap. A fixed 75th-percentile rule does not by itself guarantee this condition or identify the true support size. The theoretical noise bound $\tau_n$ is distinct from the extraction quantile $\tau=0.75$.

Under these local-surrogate, compatibility and score-separation assumptions, Proposition S1 yields task-specific supports \(\mathcal S_{\ell,r}\). Eq.~\eqref{eq:score_support_equivalence_lgamma} then shows that ablation and Taylor profiling recover those supports by thresholding. The task signature of token \(q\) is
\begin{equation}
\mathcal R_q=\{r:q\in\mathcal S_{\ell,r}\}.
\end{equation}
For a task set \(c\subseteq\{1,\ldots,R\}\), the retained functional-token subnetwork is the support union
\begin{equation}
U_c=\bigcup_{r\in c}\mathcal S_{\ell,r}
=\{q:\mathcal R_q\cap c\ne\varnothing\}.
\end{equation}
If \(r\in c\), then \(\mathcal S_{\ell,r}\subseteq U_c\). Let \(\xi_t^{(U_c)}(x)\) denote the remainder for task \(t\) after masking. Provided the local approximation remains valid after the simultaneous mask is applied to all analyzed matrices and \(\xi_r^{(U_c)}(x)\approx\xi_r(x)\), the surrogate gives
\begin{equation}
f_r^{(U_c)}(x)
\approx
\sum_{q\in \mathcal S_{\ell,r}}z_q(x)b_{qr}+\xi_r^{(U_c)}(x)
\approx
f_r(x),
\end{equation}
which is a conditional output-preservation statement, not a general sufficiency guarantee for a nonlinear network. Simultaneous removal of many weights can change activations and invalidate a single-coordinate approximation. Preserving thresholded predictions additionally requires the joint-mask output error to remain smaller than the original distance of each output from its decision threshold (0.5 for binary outputs). Support inclusion alone does not establish this stability or margin condition. If \(s\notin c\), and the same approximation remains valid, then
\begin{equation}
f_s^{(U_c)}(x)
\approx
\sum_{q\in U_c\cap\mathcal S_{\ell,s}}z_q(x)b_{qs}+\xi_s^{(U_c)}(x).
\end{equation}
A small intersection \(U_c\cap\mathcal S_{\ell,s}\) alone does not imply reduced off-target accuracy: the magnitudes of omitted contributions, biases, approximation error and classification margins also matter. An accuracy decrease requires output changes that cross decision boundaries in directions that turn correct predictions into errors, with a net loss over any corrected errors. Sufficiency and specificity of the actual joint masks are therefore assessed empirically. Finally,
\begin{equation}
U_{c_1\cup c_2}=U_{c_1}\cup U_{c_2},
\end{equation}
which explains why functional-token combinations can be composed by union of their recovered supports. This is the mathematical counterpart of the experimental masking rule: retain scalar weight tokens important for any selected task and remove the rest.

\subsection{Why the endpoint \texorpdfstring{\(\gamma=1\)}{gamma = 1} is included and why \texorpdfstring{\(\gamma>1\)}{gamma > 1} differs}

The endpoint \(\gamma=1\) is included because the Lasso penalty has a subgradient interval at zero:
\begin{equation}
\partial |b|\big|_{b=0}=[-1,1].
\end{equation}
This interval is exactly what allows the inactive KKT condition in Eq.~\eqref{eq:kkt_inactive_lasso_lgamma} to hold with \(\widehat\beta_j=0\). Thus \(\gamma=1\) can produce exact zeros and support recovery under the irrepresentable and beta-min conditions above.

For \(0<\gamma<1\), the derivative
\begin{equation}
\frac{d}{db}|b|^\gamma=\gamma |b|^{\gamma-1}\operatorname{sign}(b),
\qquad b\ne0,
\end{equation}
is singular as \(b\to0\). This singularity makes zero locally stable against sufficiently small inactive perturbations, which is why the nonconvex proof uses local domination by \(\sum_j|v_j|^\gamma\) rather than a convex KKT certificate.

For \(\gamma>1\), neither mechanism is present. There is no Lasso subgradient interval at zero, and the derivative is finite at zero; for \(\gamma=2\), it is exactly zero at zero. Small empirical correlations can then be represented by small nonzero coefficients instead of being thresholded to exact zeros. Consequently, task supports need not have a stable active/inactive gap, and ablation or Taylor score distributions may not admit a clean threshold separating functional-token subnetworks.

\subsection{Conclusion of the argument}

Combining the pieces gives the conditional chain
\begin{equation}
\begin{aligned}
L_\gamma\ (0<\gamma\le1)
&\quad\Longrightarrow\quad
\text{support recovery in a compatible local surrogate}
\\[-1pt]
&\quad\Longrightarrow\quad
\mathcal S_{\ell,r}
\quad\Longrightarrow\quad
\widehat{\mathcal S}^{\mathrm{abl}}_{\ell,r}
=
\widehat{\mathcal S}^{\mathrm{tay}}_{\ell,r}
=
\mathcal S_{\ell,r},
\end{aligned}
\end{equation}
provided the restricted-design, beta-min, noise, curvature, local-linearization, and score-margin assumptions hold. This proves the claimed relationship between \(L_\gamma\)-induced sparse supports and threshold-separable ablation/Taylor functional-token subnetworks, while making explicit the assumptions under which the conclusion is valid.

\section{Model Architectures and Training Details}
\label{appendix:architecture}

\subsection{Multitask MLP}

The robustness simulations use a fully connected Torch MLP with three hidden layers, each containing 300 ReLU units. The input dimension is $d_{\mathrm{in}}=2P=14$ for the binary arithmetic and logic tasks, whereas paired MNIST images are flattened and concatenated to give $d_{\mathrm{in}}=2\times28\times28=1568$. The output dimension is $d_{\mathrm{out}}=RP$, with $R=5$ jointly trained rules and $P=7$ output bits per rule.

Each input sample contains two vectors, $\mathbf{x}$ and $\mathbf{y}$, concatenated into a shared input representation. The model is trained on rule indices 0--4, displayed in the paper as rules 1--5. Each rule defines a distinct transformation of the same input pair and produces an output vector of length $P$. The rule-specific outputs are concatenated and predicted by a shared network.

All tasks are introduced at the beginning of training and optimized jointly. We do not use curriculum learning, sequential task training, or task-specific pretraining. The training objective is
\[
\mathcal{L}=\frac{1}{R}\sum_{r=1}^{R}W_r\mathcal{L}_{r}+l_{\mathrm{train}}\Omega_\gamma(\theta),\qquad
\Omega_\gamma(\theta)=\sum_{B\in\mathcal{B}}\frac{1}{|B|}\sum_{q\in B}(|\theta_q|+\varepsilon)^\gamma,
\]
where $\mathcal{L}_{r}$ is the mean-squared error averaged over samples and the $P$ outputs of rule $r$, $W_r=1$ by default, $\mathcal{B}$ is the set of trainable non-bias weight matrices, and $|B|$ is the number of scalar weights in matrix $B$. The implementation uses $\varepsilon=10^{-8}$ and sums the per-matrix means rather than an unnormalized sum over all weights. Biases are excluded from the penalty. The task-weighting experiment sets one $W_r$ to 5 while retaining the factor $1/R$ and the same $l_{\mathrm{train}}$ (Supplementary Note~\ref{appendix:task_weighting_allocation}).

\paragraph{Robustness-simulation setup.}
\begin{itemize}
    \item Optimizer: Adam with learning rate $0.001$.
    \item Batch size: 128.
    \item Hidden layers: three layers with 300 units per layer.
    \item Dataset size: 5{,}000 samples with a 50/50 training--validation split.
    \item Independent repeats: 30, starting from seed 300.
    \item Number of jointly trained tasks: 5 in the default experiments.
    \item Regularization exponent: $\gamma=1$ in the robustness simulations; $\gamma\in\{0.5,1,2\}$ in the regularizer-form comparison.
    \item Sparsity coefficient: $l_{\mathrm{train}}=5.0$ for MLP robustness simulations.
\end{itemize}

\paragraph{Training configurations by experiment.}
Saved training configurations distinguish the experiments listed below. All listed configurations use Adam, 5,000 training epochs, rules 1--5 (implementation indices 0--4), and a checkpoint interval of 1,000 epochs. Each experimental condition has 30 analyzed independent runs, except the TinyTransformer robustness analysis, which includes 23. The sweep contains 21 conditions; the task-weighting experiment contains six. The four mother-network conditions used for recomposition are drawn from the sweep and inherit its training settings.
\begin{center}
\small
\begin{tabular}{p{0.49\linewidth}rr}
\toprule
Experiment & Learning rate & Training batch size \\
\midrule
MLP robustness: binary, logic and MNIST & $10^{-3}$ & 128 \\
TinyTransformer robustness & $10^{-3}$ & 128 \\
Sparsity-strength/exponent sweep (Figs.~\ref{fig:main4}, \ref{fig:main5}) & $10^{-4}$ & 512 \\
Task weighting (Fig.~\ref{fig:main6}) & $5\times10^{-2}$ & 128 \\
Recomposition mother networks (Fig.~\ref{fig:main7}b,c) & $10^{-4}$ & 512 \\
\bottomrule
\end{tabular}
\end{center}
The TinyTransformer analysis uses the 23 saved runs trained with batch size 128. The seven runs trained with batch size 500 are excluded to keep the training batch size uniform. Training batch sizes in this table are distinct from the analysis batch size of 64 specified in the reproduction protocol.

\subsection{TinyTransformer}

The TinyTransformer provides a lightweight non-MLP architecture for testing whether functional-token extraction depends on fully connected hidden layers. Each paired binary input is split into left and right halves, and both halves are projected into a shared 32-dimensional token space. The resulting two-token sequence is processed by two encoder-style self-attention blocks with feedforward width 64. The final representation is obtained by mean-pooling the two tokens and passing the pooled vector through a linear prediction head with $RP$ outputs.

In the robustness simulation, the TinyTransformer is trained on the same Binary Sequence Calculation rules as the MLP, using rule indices 0--4. The encoder is trainable (\texttt{train\_encoder=True}), so both the token encoder and output head are optimized jointly. Functional tokens are constructed from scalar weight coordinates in the trained parameter matrices, profiled and retained with the same definitions used for the MLP experiments.

\paragraph{Robustness-simulation setup.}
\begin{itemize}
    \item Optimizer: Adam with learning rate $0.001$.
    \item Training batch size: 128; 23 independent runs included in the analysis.
    \item Token/hidden dimension: 32.
    \item Transformer encoder blocks: two.
    \item Feedforward hidden dimension: 64.
    \item Dataset size: 5{,}000 binary-pair samples with a 50/50 training--validation split.
    \item Number of jointly trained arithmetic tasks: 5.
    \item Regularization exponent: $\gamma=1$.
    \item Sparsity coefficient: $l_{\mathrm{train}}=0.5$ for the transformer robustness simulation.
\end{itemize}

\section{Ablation and Taylor Profiling}
\label{appendix:Taylor}

All functional-token analyses in the main text use scalar weight coordinates as tokens. For a trained model with parameters $\theta$ and rule-specific output block $f_r$, ablating a weight token $q$ means replacing $\theta_q$ by zero while leaving all other parameters fixed. The ablation importance score is
\begin{equation}
S^{\mathrm{abl}}_{r,q}=\mathbb{E}_{x\sim\mathcal{D}_r}\left\|f_r(x;\theta)-f_r(x;\theta_{-q})\right\|_2^2,
\end{equation}
where $\theta_{-q}$ denotes the parameter vector after zeroing coordinate $q$.

As a complementary first-order approximation, we estimate the squared output sensitivity of each scalar weight:
\begin{equation}
S^{\mathrm{tay}}_{r,q}=\theta_q^2\,\mathbb{E}_x\left\|\frac{\partial f_r(x;\theta)}{\partial\theta_q}\right\|_2^2.
\end{equation}
This approximates the squared first-order output perturbation associated with removing a scalar weight; it is not a task-loss gradient attribution. For each batch $B$, rule $r$ and probe $k=1,\ldots,K$, independent Rademacher signs $z_{i,d}^{B,k,r}\in\{-1,+1\}$ are drawn for each sample $i$ and output coordinate $d$. The estimator is
\begin{equation}
g_{B,k,r,q}=\frac{\partial}{\partial\theta_q}\sum_{i\in B}\sum_{d=1}^{P}z_{i,d}^{B,k,r}f_{r,d}(x_i;\theta),\qquad
\widehat S_{r,q}=\frac{1}{NK}\sum_B\sum_{k=1}^{K}(\theta_q g_{B,k,r,q})^2.
\end{equation}
Here $N$ is the total number of profiled samples. The expectation over the probes equals $\theta_q^2N^{-1}\sum_i\|\partial f_r(x_i)/\partial\theta_q\|_2^2$. There is no additional division by the output dimension $P$. The Taylor runs in Figs.~\ref{fig:main3}c,d,f and~\ref{fig:main5}a used $K=4$.

For multiplicative gates $m_q$, $\partial f_r/\partial m_q|_{m=1}=\theta_q\partial f_r/\partial\theta_q$; thus this scalar-weight score corresponds to the squared gate-Jacobian quantity in the local theoretical model.

\subsection{Token-Extraction and Modularity Settings}

The extraction quantile is fixed at $\tau=0.75$. For each trained network $n$, profiled weight matrix $b$ and rule $r$, the numerical threshold is
\begin{equation}
T_{n,b,r}=\operatorname{Quantile}_{0.75}\{S_{n,b,r,q}:q\in\mathcal C_{n,b},\ S_{n,b,r,q}\text{ finite}\},
\end{equation}
where $\mathcal C_{n,b}$ is the set of profiled candidate coordinates. The important mask is
\begin{equation}
I_{n,b,r,q}=\mathbf{1}\{S_{n,b,r,q}\text{ finite}\}\mathbf{1}\{S_{n,b,r,q}>\epsilon_S\}\mathbf{1}\{S_{n,b,r,q}\ge T_{n,b,r}\}.
\end{equation}
The quantile is computed before score-floor filtering, so finite zero scores enter its calculation. The numerical threshold is recomputed for every network, matrix and rule, and at every profiled training snapshot. Fixed $\tau$ is not a fixed numerical score threshold. For a rule set $C$, masks are combined by union,
\begin{equation}
I_{C,b,q}=\max_{r\in C}I_{b,r,q}.
\end{equation}
We construct a mask separately for each profiled weight matrix and apply all masks simultaneously, including the output weights. Biases and other parameters outside the masking set remain unchanged. This subnetwork intervention differs from zeroing one scalar weight at a time to compute its importance. The retained-weight ratio is
\begin{equation}
\rho_{\mathrm{mask}}(C)=\frac{\sum_b\sum_{q\in W_b}I_{C,b,q}}{\sum_b|W_b|}.
\end{equation}
Its numerator counts mask-selected positions; its denominator counts all scalar coordinates in the analyzed weight matrices, including zero-valued entries and excluding biases. It is not the nonzero fraction of the full network.

\paragraph{Configuration provenance and reproduction checks.}
The saved summaries for Figs.~\ref{fig:main1}c--f and~\ref{fig:main3}a,b,d,f record $\tau=0.75$, $\epsilon_S=10^{-9}$, $\epsilon_{\mathrm{out}}=10^{-8}$ and no candidate cap; the score floor is also present in all 150 original analysis exports. The current figures use 143 of these exports: 120 MLP exports and 23 TinyTransformer exports. Seven TinyTransformer runs trained with batch size 500 are excluded from Fig.~\ref{fig:main3}f, leaving the 23 runs trained with batch size 128; all original exports are preserved. The Taylor groups record four probes.

For reproducibility, we specify the current analysis configuration for Figs.~\ref{fig:main4} and~\ref{fig:main5}a: $\tau=0.75$, $\epsilon_S=10^{-9}$, all scalar-weight candidates, all available test samples and an analysis batch size of 64. Fig.~\ref{fig:main4} uses ablation profiling with $\epsilon_{\mathrm{out}}=10^{-24}$; Fig.~\ref{fig:main5}a uses randomized Taylor profiling with $\epsilon_{\mathrm{out}}=10^{-8}$, four probes and seed 0. The verified execution devices are CUDA for ablation and CPU for Taylor; fixing the Taylor device preserves the random-probe sequence. These settings define the adopted reproduction protocol.

The Fig.~\ref{fig:main4}c,d distributions match the saved importance-weighted NIP results and implement Eq.~\eqref{eq:modularity}; rule-count purity is a different statistic and was not used for these panels. Recomputing six representative ablation checkpoints with the adopted configuration gave a maximum absolute difference in $M$ of $1.04\times10^{-7}$, identical active-token counts, and identical single-rule relative accuracies and retained-weight ratios. Recomputing all 600 Taylor checkpoints underlying Fig.~\ref{fig:main5}a gave a maximum absolute difference in $M$ of $1.60\times10^{-8}$ and identical active-token counts. These checks reassess saved trained models rather than retrain them.

For Fig.~\ref{fig:main6}, all 180 newly trained 300-unit models underwent fresh scalar-weight ablation profiling. Fig.~\ref{fig:main7}b,c uses fresh profiling of all 120 saved 300-unit mother models, and Fig.~\ref{fig:main7}d uses fresh profiling of three saved auxiliary joint models. All these analyses use $\tau=0.75$, $\epsilon_S=10^{-9}$, $\epsilon_{\mathrm{out}}=10^{-8}$, perturbation batch size 64 and no candidate cap. All 194,700 scalar weights, including the output matrix, are candidates; masks are constructed separately per matrix and applied simultaneously, while biases remain unchanged. Fig.~\ref{fig:main6} uses $l_{\mathrm{train}}=5$ and $\gamma=1$, without display rescaling. The recomposition conditions retain the sweep's experiment/display strength mapping. The saved $10^{-6}$ score-ratio field is not a selection threshold in the \texttt{score\_quantile} branch, and probe-count metadata in ablation exports do not imply Taylor profiling.

Training regularization strength $l_{\mathrm{train}}$, extraction quantile $\tau$, realized score threshold $T$, score floor $\epsilon_S$ and output-energy constant $\epsilon_{\mathrm{out}}$ have distinct roles. In the saved Lasso/exponent sweep used for Figs.~\ref{fig:main4} and~\ref{fig:main5}, the displayed strength is $l=5l_{\mathrm{train}}$. Thus displayed $l=0,1,5,10$ corresponds to $l_{\mathrm{train}}=0,0.2,1,2$. Experiment-specific training settings are provided in Supplementary Note~\ref{appendix:architecture}; task-weighting and recomposition/cost protocols are detailed in Supplementary Notes~\ref{appendix:task_weighting_allocation} and~\ref{appendix:recomposition}, respectively. Rule accuracy is mean agreement over samples and output bits after thresholding predictions at $>0.5$, not exact-vector accuracy; relative accuracy divides this measure by that of the full network. In the main-text definition of $M$, active tokens satisfy $a_q>\epsilon_M=10^{-12}$, distinct from the raw-score floor $\epsilon_S$; $M$ is undefined if no token is active.

\section{Implementation Details}
\label{appendix:implementation}

Token-level analyses are performed from saved PyTorch activations, model weights, and validation data, with MATLAB-compatible exports retained only for figure-generation compatibility.

\begin{itemize}
    \item Export activations, outputs, task labels, and non-bias trainable weight matrices from trained models.
    \item Compute ablation scores $S^{\mathrm{abl}}_{r,q}$ or Taylor scores $S^{\mathrm{tay}}_{r,q}$ for scalar weight tokens.
    \item Convert scores into rule-wise important masks using the fixed score-quantile rule in Eq.~\ref{eq:important_mask}.
    \item Construct each retained rule combination by taking the union of scalar weight tokens important to any selected rule.
    \item Construct the mask separately for each profiled matrix, then simultaneously zero unselected coordinates across all profiled matrices, including output weights; biases and other unmasked parameters remain unchanged.
    \item Compute relative accuracy matrices for single-rule functional tokens and multi-rule token combinations.
\end{itemize}

\section{Task Family Definitions}
\label{appendix:tasks}

This section summarizes the task families used in the robustness simulations. The implementation uses zero-based rule indices, whereas the paper displays the same rules as rule~1, rule~2, and so on. The default active rule set is indices 0--4, corresponding to rules 1--5.

\subsection{Binary Sequence Calculation}
\label{appendix:binary_sequence}

This task family applies arithmetic rules to pairs of binary input patterns.

\paragraph{Input.}
Two binary vectors $x,y \in \{0,1\}^{P}$ are sampled as paired binary patterns. In the robustness simulation, $P=7$, 5{,}000 unique pairs are sampled without replacement from the $2^P\times2^P$ possible pairs, and the model receives the concatenated vector $[x;y]\in\{0,1\}^{2P}$.

\paragraph{Output.}
Each selected rule produces a binary vector of length $P$. With $R$ selected rules, the outputs are concatenated into
\[
Y \in \{0,1\}^{RP}.
\]

\paragraph{Rule Set.}
Let $a=\mathrm{bin2dec}(x)$ and $b=\mathrm{bin2dec}(y)$; $a$ and $b$ are the arithmetic operands used in the main text and figures, whereas $x$ and $y$ denote their binary encodings. Arithmetic rules are applied in integer space and then mapped back to binary using $\mathrm{dec2bin}$, retaining only the last $P$ bits:
\begin{enumerate}
    \item Addition: $a+b$
    \item Multiplication: $ab$
    \item Square of $a$: $a^2$
    \item Square of $b$: $b^2$
    \item Mixed multiplication: $a(a+b)$
    \item Alternative interaction: $b(a+b)$
\end{enumerate}

\subsection{Sparse Pattern Logic}
\label{appendix:pattern_sequence}

This task family applies elementwise logical operations to sparse binary patterns.

\paragraph{Input.}
Two binary vectors $x,y\in\{0,1\}^{P}$ are sampled independently from a sparse Bernoulli distribution. In the robustness simulation, $P=7$ and the sparsity parameter is 0.7, so each bit is zero with probability 0.7 and one with probability 0.3. The model receives the concatenated vector $[x;y]\in\{0,1\}^{2P}$. Sample rows are shuffled and split without deduplicating pattern pairs across the training and held-out sets. Each of the 30 saved runs contains 2,500 held-out pairs.

\paragraph{Output.}
Each selected rule produces a binary vector in $\{0,1\}^{P}$. With $R$ selected rules, the final output is
\[
Y \in \{0,1\}^{RP}.
\]

\paragraph{Rule Set.}
Logical operations are applied elementwise:
\begin{enumerate}
    \item AND: $x\wedge y$
    \item OR: $x\vee y$
    \item XOR: $x\oplus y$
    \item NOT of the first input: $\neg x$
    \item Identity of the first input: $x$
    \item Identity of the second input: $y$
\end{enumerate}

\subsection{MNIST Digit-Pair Arithmetic}
\label{appendix:mnist_pairs}

This task family applies arithmetic rules to MNIST digit labels while presenting the corresponding image pairs as input.

\paragraph{Input.}
Two MNIST images $I_1,I_2$ with digit labels $d_1,d_2\in\{0,\dots,9\}$ are selected. In the robustness simulation, digit pairs are balanced over the 100 possible label pairs, and image instances for each digit are sampled from the MNIST training split. Images are normalized, flattened, and concatenated into a single input vector. The split is performed on generated image-pair rows rather than on unique source images, so the same source image can occur in both sets. Each of the 30 saved runs contains 2,500 held-out pairs.

\paragraph{Output.}
Rules operate on integer labels $d_1,d_2$. Each rule output is converted to binary and truncated to the last $P=7$ bits. With $R$ selected rules, the output is
\[
Y \in \{0,1\}^{RP}.
\]

\paragraph{Rule Set.}
\begin{enumerate}
    \item Addition: $d_1+d_2$
    \item Multiplication: $d_1d_2$
    \item Square of $d_1$: $d_1^2$
    \item Square of $d_2$: $d_2^2$
    \item Mixed multiplication: $d_1(d_1+d_2)$
    \item Alternative interaction: $d_2(d_1+d_2)$
\end{enumerate}

\section{Representational Analysis}
\label{appendix:representational}

We characterized the internal representations of the trained networks using three complementary measures: cross-layer representational similarity, within-layer spectral complexity, and pairwise hidden-unit correlation.

\paragraph{Cross-layer representational similarity.}
We quantified representation similarity across layers using linear centered kernel alignment (CKA). For each layer $\ell$, hidden-unit activations were collected across the same set of input samples and arranged into an activation matrix $\mathbf{X}^{(\ell)}$. After centering the activation matrices across samples, linear CKA between layers $\ell$ and $m$ was computed as

\[
\mathrm{CKA}(\ell,m)
=
\frac{
\left\|
\tilde{\mathbf{X}}^{(\ell)\top}
\tilde{\mathbf{X}}^{(m)}
\right\|_F^2
}{
\left\|
\tilde{\mathbf{X}}^{(\ell)\top}
\tilde{\mathbf{X}}^{(\ell)}
\right\|_F
\left\|
\tilde{\mathbf{X}}^{(m)\top}
\tilde{\mathbf{X}}^{(m)}
\right\|_F
}.
\]

Higher CKA values indicate greater similarity between the representational geometries of two layers.

\paragraph{PCA spectral entropy.}
To characterize the complexity of the representation within each layer, we performed principal component analysis on its activation matrix. Let $\lambda_i^{(\ell)}$ denote the fraction of variance explained by the $i$-th principal component of layer $\ell$, normalized such that

\[
\sum_i \lambda_i^{(\ell)} = 1.
\]

The PCA spectral entropy was then defined as

\[
\mathcal{H}^{(\ell)}
=
-\sum_i
\lambda_i^{(\ell)}
\log \lambda_i^{(\ell)}.
\]

A low spectral entropy indicates that the representation is concentrated in a small number of principal directions, whereas a high value indicates that variance is distributed more broadly across representational dimensions.

\paragraph{Pairwise hidden-unit correlation.}
To quantify statistical dependence among hidden units within each hidden layer, we computed pairwise Pearson correlations between their pre-activation responses. Let

\[
\mathbf{H}^{(\ell)}
=
\left[
\mathbf{h}^{(\ell)}_1,
\ldots,
\mathbf{h}^{(\ell)}_{d_\ell}
\right]
\in
\mathbb{R}^{N\times d_\ell}
\]

denote the pre-activation response matrix of layer $\ell$, where $N$ is the number of input samples and $d_\ell$ is the number of hidden units. The correlation between hidden units $i$ and $j$ was calculated as

\[
\rho_{ij}^{(\ell)}
=
\frac{
\sum_{n=1}^{N}
\left(h_{ni}^{(\ell)}-\bar{h}_{i}^{(\ell)}\right)
\left(h_{nj}^{(\ell)}-\bar{h}_{j}^{(\ell)}\right)
}{
\sqrt{
\sum_{n=1}^{N}
\left(h_{ni}^{(\ell)}-\bar{h}_{i}^{(\ell)}\right)^2
}
\sqrt{
\sum_{n=1}^{N}
\left(h_{nj}^{(\ell)}-\bar{h}_{j}^{(\ell)}\right)^2
}
}.
\]

We excluded the diagonal entries and retained one value for each unordered hidden-unit pair by extracting the upper triangular portion of the correlation matrix. Because both positive and negative correlations indicate statistical coupling, we analyzed the absolute correlation magnitude,

\[
c_{ij}^{(\ell)}
=
\left|
\rho_{ij}^{(\ell)}
\right|.
\]

The values were grouped into 100 equally spaced bins over $[0,1]$, and the histogram was normalized to represent the probability distribution of pairwise correlations within each layer. Distributions concentrated near zero indicate weakly correlated and more diverse hidden-unit responses, whereas greater mass at high correlation values indicates stronger coordination or redundancy among hidden units.

\section{Task-Weighting Functional Allocation Analysis}
\label{appendix:task_weighting_allocation}

This section defines the analysis in Fig.~\ref{fig:main6}, generated with the historical \texttt{example/fig56\_protocol.py} workflow and plotted with \texttt{example/figure6.py}. The accompanying Git repository provides \texttt{reproduce.py --all} to rerun the experiments and \texttt{reproduce.py --from-results} to reconstruct the figures from the supplied observations. All six conditions use the same MLP architecture with three hidden layers of 300 units. Each condition contains $K=30$ runs with seeds 300--329, giving 180 newly trained networks. Training uses 5,000 epochs, Adam with learning rate 0.05, batch size 128, $l_{\mathrm{train}}=5$ and $\gamma=1$. In Fig.~\ref{fig:main6} and its task-weighting objective, $l=l_{\mathrm{train}}$. Each seed generates 5,000 unique binary input pairs, split equally into training and held-out sets and shared across conditions. The model initialization and optimizer random seeds are the run seed plus 100 and 200, respectively. Figure labels use rules~1--5; the implementation indexes them as 0--4.

\paragraph{Experimental design.}
Let $c=0$ denote the equal-weight control, for which $W^{(0)}_r=1$ for every rule. In intervention condition $c=u$, rule $u$ receives five-fold loss weight:
\begin{equation}
W^{(u)}_r=
\begin{cases}
5, & r=u, \\
1, & r\ne u .
\end{cases}
\end{equation}
Functional tokens are extracted independently from each trained network by ablation profiling and the fixed rule-wise score-quantile mask used in the main analysis. Importance profiling and evaluation of the masked subnetworks use the same held-out set, as in the robustness and recomposition analyses; reported retention therefore measures preservation on that analysis set, rather than generalization to a second untouched evaluation set.

\paragraph{Condition-level matrices.}
Let $A^{\mathrm{full}}_{c,r,k}$ be the unmasked-network accuracy in condition $c$, evaluated on rule $r$ in run $k$, and let $\rho_{c,r,k}$ be the retained scalar weight fraction of the functional token defined for rule $r$. Bars denote averages over runs. Panels a and b report
\begin{equation}
\Delta A_{u,r}=\overline{A}^{\mathrm{full}}_{u,r}-\overline{A}^{\mathrm{full}}_{0,r},
\qquad
\Delta\rho_{u,r}=\overline{\rho}_{u,r}-\overline{\rho}_{0,r},
\end{equation}
for every upweighted rule $u$ and evaluated/token-defining rule $r$; panel b displays $10^4\Delta\rho_{u,r}$.

To define panel c, let $A^{\mathrm{ret}}_{c,s\mid r,k}$ be the relative accuracy on evaluated rule $s$ when only the functional token defined for rule $r$ is retained in run $k$. Its specialization margin is
\begin{equation}
m_{c,r,k}
=
A^{\mathrm{ret}}_{c,r\mid r,k}
-
\frac{1}{R-1}\sum_{s\ne r}A^{\mathrm{ret}}_{c,s\mid r,k}.
\end{equation}
Panel c shows $\overline{m}_{u,r}$ for each upweighted condition and rule-defined token. Unlike panels a and b, this panel reports the margin in the intervention condition itself rather than a control-subtracted change.

\paragraph{Target-rule summaries and paired trials.}
Panel d uses the five diagonal condition-level pairs $(\Delta A_{u,u},10^4\Delta\rho_{u,u})$. For a metric $X$ (accuracy or retained-weight ratio), the implemented marginal error is $e_{c,X}=\sigma_{c,X}/\sqrt K$, where $\sigma_{c,X}^2=K^{-1}\sum_k(X_{c,k}-\overline X_c)^2$ (population standard deviation, \texttt{ddof=0}). The displayed error for a difference is $e_{\Delta X}=\sqrt{e_{u,X}^2+e_{0,X}^2}$; the weight-ratio error is multiplied by $10^4$ on the plotted axis. The combined marginal uncertainty omits the covariance between paired observations. The displayed Pearson coefficient is descriptive across the five rule means.

For panel e, upweighted and control runs are matched by repeat index. The target-rule differences are
\begin{equation}
\delta A_{u,k}
=
A^{\mathrm{full}}_{u,u,k}-A^{\mathrm{full}}_{0,u,k},
\qquad
\delta\rho_{u,k}
=
\rho_{u,u,k}-\rho_{0,u,k}.
\end{equation}
Pooling the five rules gives $RK=150$ matched pairs. The figure reports their Pearson correlation ($r=0.7835$), nominal two-sided $p$ value ($p=2.18\times10^{-32}$), and least-squares fit. Conditions share run seeds and the same control networks; this pooled $p$ value is not adjusted for clustering by seed, shared controls, or task identity.

\section{Compositional Reconstruction of Functional Tokens}
\label{appendix:recomposition}

\subsection{Source networks and component functional tokens}

For each sparsity condition, a mother MLP was first trained jointly on rules 1--5 of the Binary Sequence Calculation task family under $L_{\gamma}$ regularization, with $\gamma=1$. The sparsity-strength parameter $l$ therefore denotes the Lasso strength used during training of the mother network, rather than a regularization term applied during recomposition.

For each of four displayed regularization strengths and 30 independent trials, the saved 300-unit mother model was profiled again using scalar-weight ablation and the masking procedure described above. The 120 fresh analyses produced masks for addition, multiplication, square and composite rule 5. Each mask and its learned parameters were then held fixed while evaluating the direct rule-5 control and the two recomposition pathways, giving 360 evaluations. Recomposition does not retrain or fine-tune the components, and it does not merge their masks.

Each functional token retained the architecture and learned parameter values inherited from its mother network, while parameters excluded by its functional mask were set to zero. The masks of different functional tokens were not combined through a union of scalar weight coordinates. Instead, each token remained a separate masked MLP, and recomposition was implemented by explicitly connecting the output of one token to the input of another.

No parameters were retrained or fine-tuned during this procedure, and no additional $L_1$ penalty or other optimization objective was applied after the component tokens were connected. The resulting reconstructions therefore constitute zero-shot compositions of fixed functional subnetworks.

\subsection{Notation and output selection}

Let the original input be the concatenation of two binary vectors,
\[
\mathbf{u}
=
\begin{bmatrix}
\mathbf{x}\\
\mathbf{y}
\end{bmatrix},
\qquad
\mathbf{x},\mathbf{y}\in\{0,1\}^{7}.
\]

The corresponding integer operands are $a=\operatorname{bin2dec}(\mathbf{x})$ and $b=\operatorname{bin2dec}(\mathbf{y})$, using the original task encoding.

For a functional token $F$, its masked MLP was evaluated using the same feedforward computation as the original network. Unselected scalar weights in all profiled matrices remained zero throughout the forward pass; biases and other unmasked parameters retained their trained values.

The output of the original multitask MLP consists of consecutive seven-dimensional blocks corresponding to different rules. In the implementation, dimensions $1$--$7$ correspond to addition, dimensions $8$--$14$ to multiplication, dimensions $15$--$21$ to the square rule used to compute $a^2$, and dimensions $29$--$35$ to rule~5.

Intermediate outputs passed between functional tokens were binarized using
\[
\mathcal{B}(\mathbf{z})
=
\mathbb{I}\!\left[\mathbf{z}>0.5\right],
\]
where the threshold was applied element-wise.

\subsection{Type 1: sequential addition and multiplication}

The type~1 reconstruction implements
\[
a(a+b)
\]
as a sequential composition of the addition and multiplication functional tokens.

The addition token was first evaluated on the original input:
\[
\mathbf{o}_{\mathrm{add}}
=
F_{\mathrm{add}}(\mathbf{u}).
\]

The seven-dimensional addition output was selected from dimensions $1$--$7$ and binarized:
\[
\hat{\mathbf{z}}_{\mathrm{add}}
=
\mathcal{B}
\left(
\mathbf{o}_{\mathrm{add}}^{(1:7)}
\right).
\]

This intermediate output was concatenated with the original vector $\mathbf{x}$:
\[
\mathbf{u}_{\mathrm{T1}}
=
\begin{bmatrix}
\hat{\mathbf{z}}_{\mathrm{add}}\\
\mathbf{x}
\end{bmatrix}.
\]

The resulting vector was supplied to the multiplication functional token:
\[
\mathbf{o}_{\mathrm{T1}}
=
F_{\mathrm{mult}}
\left(
\mathbf{u}_{\mathrm{T1}}
\right).
\]

Because the final component is the multiplication token, dimensions $8$--$14$ of $\mathbf{o}_{\mathrm{T1}}$ were used as the reconstructed rule-5 output. The pathway therefore implements
\[
(a+b)a=a(a+b).
\]

The addition and multiplication tokens retained their independently extracted masks and parameter values throughout the reconstruction.

\subsection{Type 2: parallel square and multiplication followed by addition}

The type~2 reconstruction uses the identity
\[
a(a+b)=a^2+ab.
\]

The square and multiplication functional tokens were first evaluated independently on the original input:
\[
\mathbf{o}_{\mathrm{sq}}
=
F_{\mathrm{sq}}(\mathbf{u}),
\qquad
\mathbf{o}_{\mathrm{mult}}
=
F_{\mathrm{mult}}(\mathbf{u}).
\]

The square-rule output corresponding to $a^2$ was selected from dimensions $15$--$21$, whereas the multiplication output corresponding to $ab$ was selected from dimensions $8$--$14$. Both intermediate outputs were binarized:
\[
\hat{\mathbf{z}}_{\mathrm{sq}}
=
\mathcal{B}
\left(
\mathbf{o}_{\mathrm{sq}}^{(15:21)}
\right),
\]
\[
\hat{\mathbf{z}}_{\mathrm{mult}}
=
\mathcal{B}
\left(
\mathbf{o}_{\mathrm{mult}}^{(8:14)}
\right).
\]

The two intermediate vectors were concatenated:
\[
\mathbf{u}_{\mathrm{T2}}
=
\begin{bmatrix}
\hat{\mathbf{z}}_{\mathrm{sq}}\\
\hat{\mathbf{z}}_{\mathrm{mult}}
\end{bmatrix},
\]
and supplied to the addition functional token:
\[
\mathbf{o}_{\mathrm{T2}}
=
F_{\mathrm{add}}
\left(
\mathbf{u}_{\mathrm{T2}}
\right).
\]

Because the final component is the addition token, dimensions $1$--$7$ of $\mathbf{o}_{\mathrm{T2}}$ were used as the reconstructed rule-5 output. This pathway implements
\[
a^2+ab=a(a+b).
\]

Thus, type~2 connects three fixed functional tokens in a parallel-then-sequential configuration without combining their parameter masks into a single network.

\subsection{Control conditions and accuracy evaluation}

For each mother network, we evaluated three conditions:
\begin{enumerate}
    \item the functional token extracted directly for rule~5, used as the non-recomposed control;
    \item the type~1 reconstruction using addition followed by multiplication;
    \item the type~2 reconstruction using square and multiplication followed by addition.
\end{enumerate}

The corresponding full, unmasked multitask MLP was used as the performance reference. For both the full network and the directly extracted rule-5 token, output dimensions $29$--$35$ were evaluated. For the type~1 reconstruction, dimensions $8$--$14$ of the final multiplication token were evaluated, whereas dimensions $1$--$7$ of the final addition token were evaluated for type~2.

All outputs were compared with the same seven-dimensional rule-5 target,
\[
f_5(a,b)=a(a+b).
\]

Final continuous predictions were binarized using a threshold of $0.5$. For condition $k$, accuracy was calculated as the mean element-wise agreement between the binary prediction and the rule-5 target across all samples:
\[
A_k
=
\frac{1}{7N}
\sum_{n=1}^{N}
\sum_{j=1}^{7}
\mathbb{I}
\left[
\hat{b}_{nj}^{(k)}
=
y_{nj}^{(5)}
\right],
\]
where $N$ is the number of evaluated samples.

Let $A_{\mathrm{full}}$ denote the rule-5 accuracy of the corresponding full mother network. Relative reconstruction accuracy was defined as
\[
A_{\mathrm{rel}}^{(k)}
=
\frac{
A_k
}{
\max(A_{\mathrm{full}},\varepsilon)
},
\qquad
\varepsilon=10^{-8}.
\]

The numerical guard matches the implementation. Across the 360 saved evaluations, $\min A_{\mathrm{full}}=0.782$ and $\min s_5\approx0.01126$, so neither denominator guard is active. The numerator and denominator are always evaluated on rule~5. The accuracies of the addition, multiplication, and square component tasks are not independently included in this metric.

\subsection{Parameter-normalized reconstruction score}

For an extracted functional token, the retained-weight fraction was calculated as
\[
s
=
\frac{
N_{\mathrm{nz}}
}{
N_{\mathrm{total}}
},
\]
where $N_{\mathrm{nz}}$ is the number of nonzero scalar-weight entries and $N_{\mathrm{total}}$ is the total number of scalar-weight entries in the corresponding masked model. In the figure-generation implementation, these quantities are summed across weight matrices; biases are excluded. The cached retained-parameter field is an alias of the retained-weight fraction.

For the directly extracted rule-5 functional token, this fraction was
\[
s_5
=
\frac{
N_{\mathrm{nz},5}
}{
N_{\mathrm{total},5}
}.
\]

The parameter-normalized reconstruction score was defined as
\[
E_w^{(k)}
=
\frac{
A_{\mathrm{rel}}^{(k)}
}{
\max(s_5,\varepsilon)
}.
\]

Here $s_5=\rho_{\mathrm{nz},5}(l)$ in the main-text notation, and $A_{\mathrm{rel}}^{(k)}=A_{C,5}$ for configuration $C=k$. The same value of $s_5$ was used as the denominator for the directly extracted rule-5 control and for both reconstructed pathways. Consequently, $E_w^{(k)}$ provides a common parameter-normalized comparison of rule-5 reconstruction performance. It should not be interpreted as the pathway-specific parameter efficiency of the type~1 or type~2 reconstruction, because the actual retained-weight fractions of their component subnetworks are not used separately in this normalization.

\subsection{Auxiliary training and extraction cost}
\label{supp:training_cost}

The auxiliary benchmark compares three jointly trained five-task models (seeds 830--832), those models followed by weight-token extraction, and 15 separately trained single-task controls (five per seed). Each network has three hidden layers of 300 units. This auxiliary experiment uses 600 epochs of SGD, learning rate 0.05, batch size 128, and 1,000 training plus 1,000 held-out samples; it is distinct from the 5,000-epoch Adam experiments. Both joint and single-task models used $l_{\mathrm{train}}=1$, $\gamma=1$, and a regularization offset of $10^{-8}$. The trained models are reused, and weight profiling is recomputed with the protocol specified above.

For each run, the training proxy is
\[
C_{\mathrm{train}}=N_{\mathrm{parameters}}N_{\mathrm{train}}N_{\mathrm{epochs}}.
\]
For scalar-weight extraction, the implemented batching proxy is
\[
C_{\mathrm{extract}}=N_{\mathrm{test}}\sum_b\left\lceil N_{\mathrm{weight},b}/B_{\mathrm{perturb}}\right\rceil,
\]
where $N_{\mathrm{weight},b}$ counts scalar weights in matrix $b$ and $B_{\mathrm{perturb}}=64$. Each joint network has 194,700 candidate weights across four matrices. Profiling performs 194,700 single-weight perturbations in 3,045 batched perturbation-forward calls, each using 1,000 held-out samples. Thus there are 194,700,000 perturbed sample evaluations, whereas the batching proxy is 3,045,000; batching does not make those two counts equivalent. The unperturbed reference evaluation and subsequent masked-network evaluations are not included in this extraction proxy.

The training proxies are $1.17381\times10^{11}$ for the joint network and $5.61621\times10^{11}$ for the five separate networks combined. The benchmark adds the differently scaled training and extraction proxies to obtain a predefined heuristic total, $1.17384045\times10^{11}$ for joint training plus extraction. This sum is neither a FLOP count nor a wall-clock cost, and does not establish a hardware-independent or measured speed-up. Each run is normalized by its own joint-training-plus-extraction total; the separate-training ratio is 4.7845. Recomposition is excluded.

Fresh weight extraction took a mean of 13.42 seconds (sample SD 0.33 seconds) across the three runs on the recorded CUDA device. These measurements describe the new extraction stage only; they are not combined with historical training times. The mean full-network accuracies were 0.7641 for joint training and 0.8677 for separate training, so the benchmark is not an equal-accuracy comparison. Extracted tokens retained a mean relative accuracy of 0.9527 (s.e.m. 0.0047), averaging over the five rules within each joint model and then over the three models.

\end{document}